\def\year{2021}\relax
\documentclass[letterpaper]{article} 
\usepackage{aaai21}  
\usepackage{times}  
\usepackage{helvet} 
\usepackage{courier}  
\usepackage[hyphens]{url}  
\usepackage{graphicx} 
\usepackage{natbib}  
\usepackage{caption} 

\usepackage{amsfonts}
\usepackage{amsmath}
\usepackage{placeins}
\usepackage{diagbox}
\usepackage{todonotes}
\usepackage{graphicx}
\usepackage{subcaption}
\usepackage{wrapfig}
\usepackage[toc,page]{appendix}
\usepackage{textcomp}
\usepackage{amsthm}
\usepackage{longtable}
\newtheorem{theorem}{Theorem}
\newtheorem{lemma}[theorem]{Lemma}
\usepackage[switch]{lineno}

\title{Deep Contextual Clinical Prediction with Reverse Distillation}

\author {
   Rohan Kodialam,\textsuperscript{\rm 1}
   Rebecca Boiarsky,\textsuperscript{\rm 1}
   Justin Lim,\textsuperscript{\rm 1} \\
   Neil Dixit,\textsuperscript{\rm 2}
   Aditya Sai,\textsuperscript{\rm 2}
   David Sontag\textsuperscript{\rm 1} \\ 
}
\affiliations {
    \textsuperscript{\rm 1}MIT CSAIL \& IMES
    \textsuperscript{\rm 2}Independence Blue Cross  \\
    kodialam@alum.mit.edu, rboiar@mit.edu, justinl@mit.edu, \\ 
    Neil.Dixit@ibx.com, Aditya.Sai@ibx.com, dsontag@csail.mit.edu
}

\begin{document}

\maketitle

\begin{abstract}

     Healthcare providers are increasingly using machine learning to predict patient outcomes to make meaningful interventions. However, despite innovations in this area, deep learning models often struggle to match performance of shallow linear models in predicting these outcomes, making it difficult to leverage such techniques in practice. In this work, motivated by the task of clinical prediction from insurance claims, we present a new technique called \emph{reverse distillation} which pretrains deep models by using high-performing linear models for initialization. We make use of the longitudinal structure of insurance claims datasets to develop Self Attention with Reverse Distillation, or SARD, an architecture that utilizes a combination of contextual embedding, temporal embedding and self-attention mechanisms and most critically is trained via reverse distillation. SARD outperforms state-of-the-art methods on multiple clinical prediction outcomes, with ablation studies revealing that reverse distillation is a primary driver of these improvements. Code is available at https://github.com/clinicalml/omop-learn.

\end{abstract}

\section{Introduction\label{intro}}
Machine learning of predictive models on health data is widely used to guide preventative, prophylactic and palliative care. 
We focus on a subset of electronic medical records that are frequently found as part of health insurance claims or as administrative data in large hospital systems. For each patient, we receive a time series of \emph{visits} -- single continuous interactions of a patient with the healthcare system -- and \emph{codes} -- the medical events occurring during each visit. These codes detail the specialties of visited doctors, diagnoses, procedures, the administration of drugs, and other medical concepts.

Several aspects of these claims data make the machine learning challenge unique from other settings where sequential data is observed (e.g., natural language processing). First, the data is extremely sparse. Second, multiple observations are recorded during a single visit (e.g., diagnoses, procedures, medications) and the vocabulary of medical concepts is often in the tens of thousands. Third, visits correspond to highly irregularly-spaced time series of events, since care is often administered in short bursts punctuated by long gaps. Variable timescales must be simultaneously accounted for, since the time between visits made by a single patient can vary from years to days. 

Deep learning suggests a path to improving predictive performance by learning representations of longitudinal health records that capture a patient's medical status and potential future risks. 
State-of-the-art models in the literature have largely focused on shorter-term prediction over horizons of days or weeks, most notably during a single hospital visit or in the immediate aftermath of a visit \cite{25}. Approaches to longer-term prediction often rely on manually feature-engineering longitudinal health data into patient state vectors \cite{4,5,6,7}, as opposed to training end-to-end from raw longitudinal EHR data. Due to this heuristic approach, these methods cannot fully exploit the temporal nature of EHR data, nor the relationships between clinical concepts. 

We introduce Self Attention with Reverse Distillation, or SARD, a self-attention based architecture for longitudinal health data, which uses a self-attention mechanism \cite{11} to extract meaning from the temporal structure of medical claims and the relationships between clinical concepts. Our architecture is inspired by BEHRT \cite{li2020behrt}, which recently outperformed previous deep learning algorithms for medical records including RETAIN \cite{14} and Deepr \cite{nguyen2016}. Building off of BEHRT, our major contribution is our novel pre-training procedure, reverse distillation (RD); our architecture also differs in several other key aspects. 

In reverse distillation, we first initialize our model to mimic a performant linear model, and subsequently fine-tune. We find empirical evidence that reverse distillation acts as an effective way to perform soft feature selection over complex feature spaces, such as multidimensional time-series data. We further establish statistically significant gains against strong baselines in terms of predictive performance for three long-term tasks -- predicting the likelihood of a patient dying, requiring surgery, and requiring hospitalization -- with clear applications to preventative and palliative healthcare. Our experiments also establish that reverse distillation is a key driver behind these wins, and pave the way for the use of this method in future research.

In summary, we present the following contributions:
\begin{itemize}
    \item SARD, a transformer architecture which uses an explicit visit representation to better encode claims data. SARD also uses a convolutional prediction head to ingest the outputs of its transformer layers, in contrast to the linear heads used in previous work.
    \item Reverse distillation, a novel and broadly applicable method of initializing machine learning models using high-performing linear models.
    \item An introspection analysis of how reverse distillation allows SARD, and deep models in general, to generalize better and make more accurate predictions by effectively regularizing deep models to make good use of features known to be clinically meaningful.
\end{itemize}


\section{Related Work \label{related}}

Many recent works analyze how deep learning can be applied to clinical prediction \cite{1,2,3,12,14,17,19,22,23}. Several approaches use recurrent neural networks (RNNs) to ingest medical records, and achieve excellent performance on tasks like predicting in-patient mortality upon hospital admission \cite{1}. Further refinements add learned imputation to account for missingness \cite{3}, and improvements in featurizing time by using architectures like bi-directional RNNs \cite{21}, explicit temporal embeddings \cite{baytas2017patient} and two-level attention mechanisms to find the influence of past visits on a prediction \cite{14,24}. Research has also focused on using convolutional neural networks (CNNs) to develop better embeddings of clinical concepts passed into a recurrent model \cite{22}, and graphically representing the patient-clinician relationship to augment health record data \cite{23}. Self-attention has also been used to develop relationships between medical features that have already been collapsed over the temporal dimension using recurrent methods \cite{ma2020concare} and to phenotype patients \cite{song2017attend}. More recently, self-attention was used in BERT for EHR, or BEHRT \cite{li2020behrt}, to simultaneously predict the likelihood of 301 conditions in future patient visits.

When making predictions with horizons of months or years, the state-of-the-art is often still simple, linear models with carefully chosen features \cite{bellamy2020evaluating, 4,5}. Recent work exploring deep-learning based approaches to long-term clinical prediction train neural networks directly on features constructed using hand-picked time windows and summary statistics \cite{6} or use denoising autoencoders to pre-process this type of data \cite{7}, and do not necessarily beat strong linear baselines \cite[Supplemental Table 1]{2}. Critically, many of these models rely on manual feature-engineering to create representations of the time-series data that forms a patient's medical record rather than learning this structure in tandem with the task at hand.

\section{SARD Model Architecture \label{architecture}}

Our model builds upon self-attention architectures \cite{11}, most recently applied in the clinical domain by the BEHRT model. SARD differs from BEHRT in several important ways. Firstly, SARD operates on visit embeddings which summarize a patient's medical events in that visit in a single input, while BEHRT encodes each diagnosis separately in a sequence, using separators to indicate the boundaries of each visit. This allows SARD to include significantly more data from a patient's history with the same computational efficiency. Secondly, SARD uses a convolutional prediction head applied to all transformed visit embeddings, while BEHRT uses dense layers applied to a single transformer output. Furthermore, BEHRT was demonstrated on a feature dimension of 301 condition codes, which did not include medications and procedures; in this paper, we apply SARD on a much larger set of 37,004 codes, spanning conditions, medications, procedures, and physician specialty. 



\begin{figure}[t]
  \centering
  \includegraphics[width=0.99\linewidth]{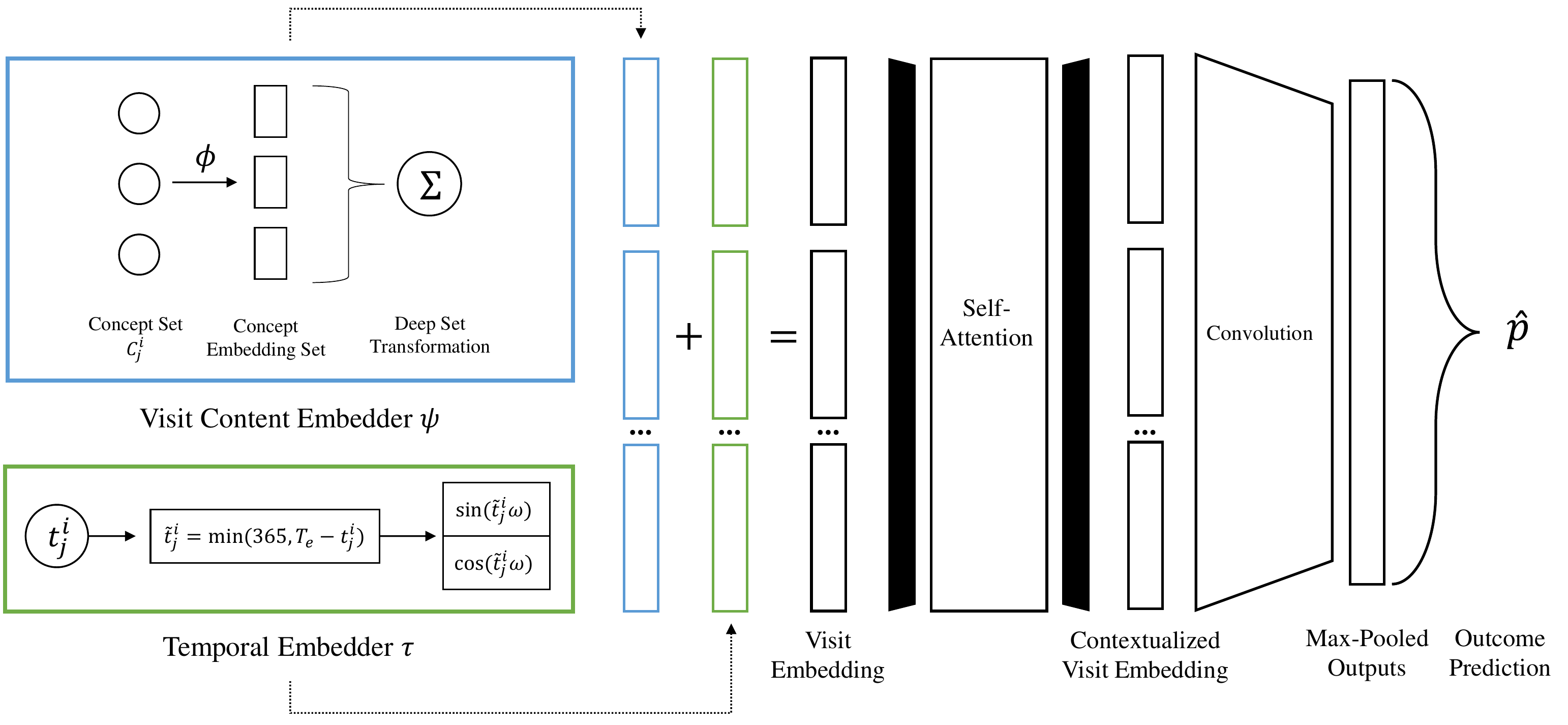}
  \caption{SARD Architecture for Longitudinal Claims Data}
  \label{figure:architecture}
\end{figure}

We use a set encoding approach to address the challenge of sparsity and the need to represent a set of data observed at each visit, and a self-attention based architecture to allow any visit's embedding to interact with another visit embedding through $O(1)$ layers, thus ensuring that we can capture temporal information and dependencies.
An overview of the architecture is provided in Figure \ref{figure:architecture}.

We denote the set of visits made by a patient $i$ by $\mathcal{V}_i$, and represent this patient's $j$th visit by $V^i_j$. We further denote the time of visit $V^i_j$ by $t^i_j$ and the set of codes assigned during visit $V^i_j$ with $C^i_j \subseteq \mathcal{C}$.

\textbf{Input Embedding}:
We adapt the method of \citet{8} to generate an initial concept embedding map $\phi:\mathcal{C}\to \mathbb{R}^{d_e}$, learned only using data in the training window to prevent label leakage. The vector representation $\psi(V^i_j) \in \mathbb{R}^{d_e}$ of each visit is calculated as $\psi(V^i_j) = \sum_{c \in C^i_j} \phi(c)$, providing invariance to permutations of the codes. This is similar to the Deep Sets paradigm, with nonlinearity provided by the embedding $\phi$ and downstream components of our architecture \cite{10}.

\textbf{Temporal Embedding}:
SARD does not explicitly encode the order of events, and visits do not occur in regular intervals. We embed the time of each visit into $\mathbb{R}^{d_e}$ using sinusoidal embeddings \cite{11}, and generate a \emph{temporal embedding} $\tau(V^i_j) = \sin(\tilde{t}^i_j \omega) || \cos(\tilde{t}^i_j \omega)$, where $\tilde{t}^i_j = \min(365,T_A-t^i_j)$ and $T_A$ represents the prediction date. This allows us to measure time relative to the prediction date. We found that clipping these relative time differences at one year increased performance -- this design choice effectively groups together all longer-term dependencies. Note that we denote concatenation with $||$, $\omega$ is a length $d_e/2$ vector of frequencies in geometric progression from $10^{-5}$ to $1$, and $\sin$ and $\cos$ are applied element-wise. 

\textbf{Self-Attention}:
We add $\psi(V^i_j)$ and $\tau(V^i_j)$ to create final encodings that represent the content and timing of visits. To contextualize visits in a patient's overall history we use multi-headed self-attention \cite{11} with $L=2$ self-attention blocks and $H=2$ heads. For efficiency, we truncate to the $n_v = 512$ most recent visits, and add padding for patients with less than $n_v$ visits, but use a masking mechanism to only allow non-pad visits to attend to each other. We apply dropout with probability $\rho_d^t=0.05$ after each self-attention block to prevent overfitting. This approach allows any visit to attend to any other, so longer-range dependencies of clinical interest can be learned.


Each layer of each head performs three affine transformations on the input embeddings, which for the first layer are $\psi(V^i_j) + \tau(V^i_j)$ for each visit $V^i_j$. These transformations produce vectors $k^i_j, q^i_j$ and $v^i_j$ respectively. We find the contextualized embedding of visit $V^i_j$ by computing raw attention weights $w^i_{j \ell} = q^i_j \cdot k^i_{\ell} / \sqrt{d_e}$, normalizing via softmax to $\tilde{w}^i_{j \ell} = \left({\sum_{r=1}^{n_v} e^{w^i_{jr}}}\right)^{-1} e^{w^i_{j \ell}}$, and taking the weighted sum $\sum_{\ell=1}^{n_v} \tilde{w}^i_{j \ell} v^i_{\ell}$. This process is then repeated at each layer using the contextualized embeddings as inputs, and residual connections are used between layers. The outputs of each head are concatenated to create final, contextualized visit representations $\tilde{\psi}(V^i_j)$.


\textbf{Convolutional Prediction Head}:
The prediction head returns an estimated probability of the target event using the outputs of the self-attention mechanism. We do so by creating $K$ convolutional kernels of size $d_e \times 1$. Then, each kernel extracts a feature from the non-pad contextualized visit embeddings by first calculating a cross-correlation versus each $\tilde{\psi}(V^i_j)$, then using a max-pooling operation to select the highest of these cross-correlations. Concatenating these outputs gives a length-$K$ real vector of extracted features. To obtain a predicted probability $\hat{p}(i)$ for each patient, we apply a sigmoid nonlinearity to this vector,  take the dot product of the transformed components with a learned vector of weights, and apply another sigmoid nonlinearity to obtain a final prediction probability.

\section{Learning with Reverse Distillation\label{rdistill}}
Reverse distillation is a novel method by which we initialize a deep model using a linear proxy. We consider a binary prediction model $f_\theta : \mathcal{X} \to [0,1]$ parametrized by $\theta$ which maps from a domain $\mathcal{X}$ of data to a probability value, and a linear model $g_w:\mathcal{X} \to [0,1] $ defined by $g_w(x) = \sigma(w^T \xi(x))$, where $\sigma$ is the sigmoid function and $\xi$ is a fixed \emph{feature engineering} transformation $\xi:\mathcal{X}\to \mathbb{R}^d$ based on heuristic domain knowledge. 

While $f_\theta$ may be a large, highly-parametrized model, $g_w$ may perform better on prediction tasks for several reasons, including the ability to select features and avoid overfitting through regularization of $w$, and the quality of the transformation $\xi$. As such, we initialize $f_\theta$ to mimic the outputs of $g_w$ in order to benefit from the structure and performance of the linear model while allowing for further data-driven improvements.



We interpret predictions $f_\theta(x)$ (resp $g_w(x)$) as indicating that the distribution of the label for data point $x$ is $\mathbf{B}(f_\theta(x))$ (resp $\mathbf{B}(g_w(x))$), where $\mathbf{B}(p)$ indicates a Bernoulli distribution with success parameter $p$. We perform reverse distillation by pre-training our deep model to optimize over $\theta$ a loss function defined by 
\begin{eqnarray}
\ell_\textrm{RD}(x) &= &- p_cg_w(x) \log f_\theta(x) \nonumber \\ 
&&- (1-g_w(x))\log(1-f_\theta(x)).
\end{eqnarray}
This algorithm is inspired by the standard knowledge distillation paradigm \cite{hinton2015distilling}, in which a simpler model is trained to mimic a complex model. To fine-tune $f_\theta$, we make use of both the true label $y(x) \in \{0,1\}$ and the prediction $g_w(x)$, combining a cross-entropy loss versus the true label
\begin{eqnarray}
\ell_\textrm{CE}(x) &= &- p_cy(x) \log f_\theta(x) \\
    &&- (1-y(x)) \log(1-f_\theta(x))
\end{eqnarray}
and the reverse distillation loss $\ell_{RD}$, to get a loss function
\begin{equation}
\ell_\textrm{tune}(x) = \ell_{CE}(x)+\alpha\ell_{RD}(x).
\end{equation}
We include a class weighting term $p_c$ equal to the ratio between the number of negative and positive training data points to encourage higher recall in our trained model, and a hyperparameter $\alpha$ to represent the weight placed on differences between $g_w(x)$ and $f_\theta(x)$. We note that cross-validation over $\alpha$ always selected 0 in our experiments, meaning that reverse distillation was only needed for initializing the model. 

{\bf Training Procedure for SARD.}
We next describe our procedure for training a SARD model with reverse distillation. All training is performed end-to-end, including the initial embedding $\phi$ of clinical concepts. 
We reverse distill from a highly $L_1$-regularized logistic regression model. As the logistic regression's predictions tend to be well-calibrated \cite{16}, we interpret its output as a distribution over outcomes. While hand-engineered features are often created for specific tasks in the clinical domain, we opt for a more general formulation. Inspired by prior work in high-performance linear models for clinical prediction \cite{4}, we construct features by aggregating codes over different temporal windows, and thus we refer to this model as a \emph{windowed} linear model. Given a time interval $W=[t_s,t_e]$, we find the feature vector corresponding to this interval for patient $i$ by finding the subset of visits 
$\mathcal{V}_i(W) = \{V^i_j\in \mathcal{V}_i | t^i_j \in W \}$
and subsequently finding the set of codes 
$\mathcal{C}_i(W) = \bigcup_{V^i_j \in \mathcal{V}_i(W)} C^i_j.$ 
We find that performance was optimized by using a multi-hot vector $f_i(W)$ of size $|\mathcal{C}|$ as the feature engineering transformation $\psi$ to map these sets of codes to real-valued vectors, with the element corresponding to concept $c \in \mathcal{C}$ set equal to $1$ if $c \in \mathcal{C}_i(W)$ and $0$ otherwise.

To capture the longitudinal nature of claims data, we use multiple windows simultaneously as features. We establish a list $\mathcal{W}_C$ of candidate windows, each of which has an end time equal to the prediction date and start times ranging from $15$ to $\infty$ days before the prediction date, as shown in Appendix Table \ref{table:hyp}. We selected the $n_W = 5$ best windows from all ${|\mathcal{W}_C| \choose n_W} $ unique window choices by comparing validation performances.


\subsection{Theoretical analysis}
We note that a deep model and a linear model making the same classifications are not necessarily learning the same classification boundary. We investigate if the self-attention model actually replicates the linear model's classification function. 

We find that it is possible to construct a set of weights such that SARD and a windowed logistic regression model have identical outputs for all inputs:
\begin{lemma}
\label{lemma1}
In the limit $d_e \to \infty, K\to \infty$ and for an appropriate choice of $\omega$, SARD can identically replicate a windowed linear model.
\end{lemma}
The proof can be found in the Appendix. The crux of the argument is that we can express a filter of the form $[[t^i_j < T]]$ for any $T$ as a linear combination of the elements $\tau(V^i_j) = sin\left(t^i_j\mathbf{\omega}\right)||cos\left(t^i_j \mathbf{\omega}\right)$, with weights determined as Fourier series coefficients. This allows SARD to replicate the windowed feature vectors of the linear model. We note that this lemma holds even with a single self-attention layer.

This result increases our confidence in our choice of architecture and its ability to generalize and improve beyond a linear model. For example, windows of the form $[[t^i_j < T]]$ implied by the linear model might be inferior to a more complex filter in the time domain. However, such filters can be learned by SARD. While the existence of this set of weights does not mean that SARD will converge to these exact weights after reverse distillation, it does highlight one possible mechanism for ensuring that the deep and linear models generalize in the same way. 

\subsection{Interpretability via Network Dissection}\label{methods:network_dissection}

We next introduce a technique to investigate whether and how reverse distillation surfaces features of the windowed linear baseline. We utilize the Network Dissection global interpretability framework of \citet{bau2018gan} to compare the outputs of the penultimate layer of SARD networks to the linear baseline's features. Our goal is to match the latent features which are inputted to the final prediction head in the deep model to the interpretable features of the linear model, as a means of both understanding which linear features are preserved using reverse distillation, as well as to aid in interpreting the deep model features which are ultimately used in prediction. To do this ``matching," we binarize the penultimate layer of the deep model by taking the sign of each output, and then calculate the Matthew's Correlation Coefficient (MCC) of each output with each windowed linear baseline feature, across all people in the test set. 

\section{Experiments}\label{experiments}

We evaluate our approach using a de-identified dataset of $121,593$ Medicare Advantage patients provided by a large health insurer in the United States. This data is mapped into the Observational Medical Outcomes Partnership (OMOP) common data model (CDM) version 6 \cite{9}. OMOP provides a normalized concept vocabulary, and although our dataset is not public, hundreds of health institutions with data in an OMOP CDM can use our code out-of-the-box to reproduce results on local datasets\footnote{https://github.com/clinicalml/omop-learn}. We also investigate the properties of reverse distillation through experimentation on synthetic data.

{\bf Baselines.} We compare to several baselines. First, we compare to the windowed $L_1$-regularized logistic regression model \cite{4} described earlier in the context of reverse distillation. Second, we compare to two of the previous state-of-the-art deep learning models for similar tasks: RETAIN \cite{14,24}, a recurrent architecture with attention, and BEHRT \cite{li2020behrt}, the transformer-based architecture which served as the jumping off point for our model. Third, we  compare to our own self-attention-based model trained without reverse distillation.


To build a BEHRT model in our data setting, we use a self-attention architecture to ingest sequences of medical codes (as in the original BEHRT model) instead of aggregated sequences of entire visits (as in SARD). This model is very similar to BEHRT, with some minor differences. Specifically, we omit the use of $SEP$ tokens and age embeddings. Due to the computational constraints imposed on both the SARD and BEHRT models, it was generally not possible to include significantly more than one year of data for a given patient, rendering a per-code age embedding superfluous. For the same computational reasons, we omit the $SEP$ token to allow more actual codes to be embedded per patient. In our initial experimentation, we found no gains from using a masked language model to pretrain transformer architectures (including both BEHRT and SARD) in our setting. We instead used the method of \citet{8} for initialization in all cases.  We further discuss our choice of baselines in the Appendix. 

We train using a single NVIDIA k80 GPU. Our algorithms are implemented in Python 3.6 and use the PyTorch autograd library \cite{pytorch}. We train our deep models using an ADAM optimizer \cite{kingma2014adam} with the hyperparameter settings of $\beta_1=0.9,\beta_2=0.98,\epsilon=10^{-9}$ and a learning rate of $\eta = 2\times 10^{-4}$. A batch size of $500$ patients was used for ADAM updates.

%
%
{\bf Prediction tasks.} We consider three tasks important for predictive healthcare:

\textbf{1}: The \emph{End of Life (EoL)} prediction task: we estimate patient mortality over a six-month window. This task is key to proactively providing palliative care to patients.

\textbf{2}: The \emph{Surgical Procedure (Surgery)} prediction task: we predict if a patient will require any surgical procedure in a six-month window. If so, an appropriate, intervention can be taken early on.

\textbf{3}: The \emph{Likelihood of Hospitalization (LoH)} prediction task: we estimate if a patient will require inpatient hospitalization in a six-month window. This allows for early interventions that could mitigate the need for hospitalization.

We split the $121,593$ patients into training, validation, and test sets of size $82,955$, $19,319$, and $19,319$ respectively. Data was collected up to the end of the calendar year 2016, and outcomes measured between April and September of 2017 -- patients who had an outcome in the three-month gap between the end of data collection and the outcome measurement were excluded from the dataset. We denote the set of all OMOP concepts used in the dataset by $\mathcal{C}$, which in our case contained $|\mathcal{C}| = 37,004$ codes. All models are trained using the SARD architecture, using reverse distillation with early stopping for both pre-training and fine-tuning. SARD models are trained with $d_e=300$ and $K=10$; we found that validation performance did not increase with larger embedding sizes or number of convolutional kernels. Early stopping and the selection of the hyperparameters as outlined in Appendix Table \ref{table:hyp} are performed using the validation set, and the parameters that maximized validation ROC-AUC are used to evaluate performance on the test set. 

Our metric for measuring the performance is the area under the receiver-operator curve (ROC-AUC), i.e. the area under a plot of the true positive rate of the model as a function of false positive rate. An equivalent interpretation is the probability that the model gives a higher score to a random positive-outcome patient than a random negative-outcome patient. Thus, ROC-AUC is a good proxy for the application of choosing which patients should receive early interventions. While in class-balanced problems metrics like accuracy are useful, and in cases of extreme class imbalance metrics like AUC-PRC may provide insights, our metric is meaningful across a wide variety of class imbalances that may occur in the clinical domain. Indeed, our class balances range from $1.8\%$ for EoL, to $8.5\%$ for LoH, to $57.8\%$ for Surgery. Nevertheless, for completeness, we also provide an AUC-PRC comparison in the Appendix, and find that SARD continues to outperform baselines.

\subsection{Main results}


\begin{table}[t]
\centering
\caption{AUC-ROC Scores on Test Set. + RD indicates that reverse distillation is used for pre-training. Increases in AUC-ROC for SARD are significant versus the closest baseline in all cases (paired $z$-test, $p<.005$).}
\label{table:auc}
\begin{tabular}{m{0.5\linewidth}m{0.1\linewidth}m{0.1\linewidth}m{0.1\linewidth}}\hline
\backslashbox{Model}{Task Name}
&{EoL}&{Surgery}&{LoH}\\\hline\hline
$L_1$-reg. logistic regression \cite{4} & 83.4 & 79.2 & 73.1\\\hline \hline
RETAIN \cite{14} &82.2 & 79.8 & 72.5\\\hline \hline
BEHRT \cite{li2020behrt} & 83.1 & 80.3 & 71.2 \\\hline
BEHRT + RD & 83.7 & 81.1 & 73.7 \\\hline \hline
SARD (no RD) & 85.0 & 82.7 & 72.7\\\hline
SARD & \textbf{85.6} & \textbf{83.1} & \textbf{74.3} \\\hline
\end{tabular}
\end{table}

As seen in Table \ref{table:auc}, our model outperforms all baselines for each of the example tasks. Increases in AUC-ROC are significant versus the closest baseline in all cases (paired $z$-test, $p<.005$) \cite{delong1988comparing}. Notably, while the SARD model has the absolute highest performance, RD pre-training still offers improvement to the BEHRT baseline; through ablation studies, we show that RD similarly improves performance across additional, varied architecture choices. In the next section, we explore the nuances of how SARD extracts clinical narratives, and qualitatively find that SARD is able to use a patient's entire medical history to contextualize visits, whereas the high-performing linear models are not able to make these connections.



{\bf Ablation Studies.}
We empirically test the design decisions made in our SARD Model Architecture section via ablation studies. These studies validate our architecture choices, as ablation of both SARD's self-attention mechanism and its convolutional prediction head lead to performance decreases. 

As seen in Figure \ref{figure:architecture}, the SARD architecture naturally splits into modular parts, the two most important of which - the transformer and the prediction head - we investigate via ablation:

\begin{itemize}
    \item \textbf{Self-attention}: 
    A key aspect of our work is its use of a self-attention architecture as a tool to ingest time-series of embedded clinical data. Until recently, RNN-based approaches \cite{1,14,21} have been the state-of-the-art, and as such we developed an ablation study in which we replace our architecture with a unidirectional recurrent GRU-cell network, leaving the rest of the network unchanged. This GRU-cell network used input dimension $d_e=300$ and hidden dimension $d_e=300$.
 In Table \ref{table:ablation}, the row \texttt{RNN (no RD)} corresponds to this ablated model trained from a random initialization, and \texttt{RNN + RD} to the ablated model trained using the same reverse distillation procedure used in SARD.
 
 To ensure that our ablation fairly compared recurrent and self-attention based approaches, we preserved all other architectural elements including the visit-level input embeddings, use of temporal embeddings (fixed-frequency sinusoidal time embeddings led to the best performance), and the prediction head to aggregate the final visit representations, which here operates on the hidden states of each element of the last layer of the RNN. We found the prediction head's aggregation to be more performant and serve as a more apt comparison than the standard recurrent technique of simply predicting from the hidden state of the last element of the last layer of the RNN. This design choice helps mitigate the fact that older visits may be `forgotten' by the RNN, by allowing these visits to directly influence the inputs of the prediction head.
 We find that the self-attention architecture is competitive with the RNN, so long as the RNN is also trained with reverse distillation. An important finding is that reverse distillation can also be used to successfully train highly-performant recurrent models, further validating the usefulness of this method and indicating that it can be used more generally. 
 
We performed a similar ablation in which we replaced the the self-attention layers with the identity, to further evaluate the value of explicitly contextualizing visits. In Table \ref{table:ablation}, the row \texttt{Identity (no RD)} corresponds to this ablated model trained from a random initialization, and \texttt{Identity + RD} to this model trained using the same reverse distillation procedure used in SARD. We find that self-attention and recurrent architectures improve performance on our surgery task, but have less of an impact on our other two tasks; why this is requires further investigation. Furthermore, the strong performance of our identity ablation speaks to the strength of our convolutional prediction head, a design choice that likely contributes to our improvement over the previous state of the art, BEHRT.

\item \textbf{Prediction Head}: 
We also ablate our convolutional prediction head by replacing it with a naive alternative which simply sums the contextualized vector representations of all visits to obtain a vector $\sum_j \tilde{\psi}(V^i_j)$ representing the entire history of patient $i$. This summed vector, which will have dimension $d_e$, is then passed into a single linear layer with sigmoid activation to make a final prediction. We use input embedding, sinusoidal time embedding and a self attention mechanism identical to those of the SARD model described in our SARD Model Architecture section. 

We find that SARD's convolutional prediction head gives performance increases when compared with this simpler alternative. Even in this regime, we again find that reverse distillation allows models to be more performant. In Table \ref{table:ablation}, the row \texttt{Summing Head (no RD)} corresponds to this ablated model trained from a random initialization, and \texttt{Summing Head + RD} to this model trained using the same reverse distillation procedure used in SARD.
\end{itemize}
 As seen in Table \ref{table:ablation}, our design choices perform as well as or better than alternatives. Importantly, our ablation studies highlight that in addition to architectural innovations, reverse distillation is a key driver in SARD's performance gains, and more generally in performance gains across diverse architectures. Indeed, the smallest difference in ablated performance was observed when SARD's self-attention architecture was replaced with a recurrent equivalent, but reverse distillation was still used for pre-training, indicating reverse distillation's universal applicability.
 
\begin{table}[t]
\centering
\caption{Ablation Study Results.  + RD indicates that reverse distillation is used for pre-training}
\label{table:ablation}
\begin{tabular}{m{0.55\linewidth}m{0.05\linewidth}m{0.1\linewidth}m{0.05\linewidth}}\hline
\backslashbox{Design Choice}{Task Name}
&{EoL}&{Surgery}&{LoH}\\\hline\hline
SARD & 85.6 & 83.1 & 74.3\\\hline
SARD (no RD) & 85.0 & 82.7 & 72.7\\\hline\hline
\multicolumn{4}{l}{Ablations Replacing Self-Attention with:}\\\hline
RNN + RD &85.5&82.8&74.1\\\hline
RNN (no RD) &84.3&82.3&72.6\\\hline
Identity + RD &85.3&81.6&74.1\\\hline
Identity (no RD) &84.3&79.9&73.2\\\hline\hline
\multicolumn{4}{l}{Ablations Replacing Convolutional Prediction Head with:}\\\hline
Summing Head + RD &84.2&82.4&74.2\\\hline
Summing Head (no RD) &83.1&81.6&72.0\\\hline

\end{tabular}
\end{table}
\subsection{Model Introspection \label{sec:introspection}}

In healthcare applications, it is critical to understand and interpret how models make predictions. In this section we employ a local, or per-prediction, method of introspecting on the SARD model; specifically, we examine which visits are most influential in the prediction head for a given individual, and how those visits leverage self-attention to contextualize. Our primary goal in this analysis is to introspect on the SARD model to better understand how its self-attention architecture leverages and transforms our input features to make improved predictions, and we note that further work would be needed before using such interpretation methods to justify clinical decisions.


To determine which visits are most influential to a prediction, we introspect directly on our convolutional prediction head. In notating this introspection, we suppress indices corresponding to batches (i.e. patients), as the introspection will be ultimately performed at the level of a single individual. 

Recall that the prediction head convolves $K$ kernels of size $d_e \times 1$ with the final contextualized visit representations, then uses a max-pooling operation to return the maximum cross-correlation between the kernel and any individual contextualized visit. For the $k^{th}$ of these $K$ kernels, denote this maximum cross-correlation value by $\chi_k$, and the maximizing visit by $\nu_i$. Let $w_k$ denote the weight given to the output from the $k^{th}$ kernel in the final linear layer mapping to a prediction. We assign a score of $s(V_j) = \sum_k [[V_j=\nu_k]] w_k \sigma(\chi_k)$ to visit $V_j$, where $\sigma$ represents the sigmoid nonlinearity applied after max-pooling. This metric represents the total importance of visit $V_j$ by summing all of its possible contributions to the final prediction.

We use these introspection techniques in the Appendix to interpret the case of a $\geq$90 year-old female patient whose death was predicted with high probability (71.1\%) by SARD, but missed by our baseline windowed linear model (5.4\% probability of death). Using the total importance metric described above, we can find the most predictive visits for our case study patient in SARD. We present her top four visits, which include visits from 2011, 2015 and 2016 in which the patient chiefly experienced cardiovascular diseases and their complications, in Appendix Table \ref{table:topshapvisits}.

We then seek to understand how each visit is contextualized by examining its attention weights in SARD's self-attention layers. For example, in our case study, we examine the visits attended to most strongly by the patient's top visit; we include these results in Appendix Table \ref{table:topattnvisits} and visualize the attention weights from her top visit in Appendix Figure \ref{fig:case_study_all4heads}. We find that while this patient's top visit occurred in 2016 and included detection of a myocardial infarction along with other cardiovascular disease, her top visit strongly attends to a cluster of visits in 2011. By carefully analyzing these visits, we find that during the 2011 visits, the patient experienced other manifestations of atherosclerotic vascular disease. We conjecture that these continued, albeit more minor, cardiovascular issues over the years provide context for the 2016 visit, and ultimately augment the risk of death associated with the events of the 2016 visit.

More generally, introspecting on the SARD model reveals that its self-attention mechanism leverages important contextual information from throughout a patient's history to gain a nuanced understanding of which parts of the medical timeline are most important for prediction. Thus, the deep model is able to make better predictions than simpler baselines when it is necessary to interpret an entire clinical narrative. In particular, in cases where SARD outperforms linear baselines, patients have significantly \emph{more} data, as measured by the patient's total number of visits, than in cases where the linear baseline outperforms (Mann-Whitney $U$ test, $p<.05$).

\begin{figure}
  \centering
  \includegraphics[width=0.9\linewidth]{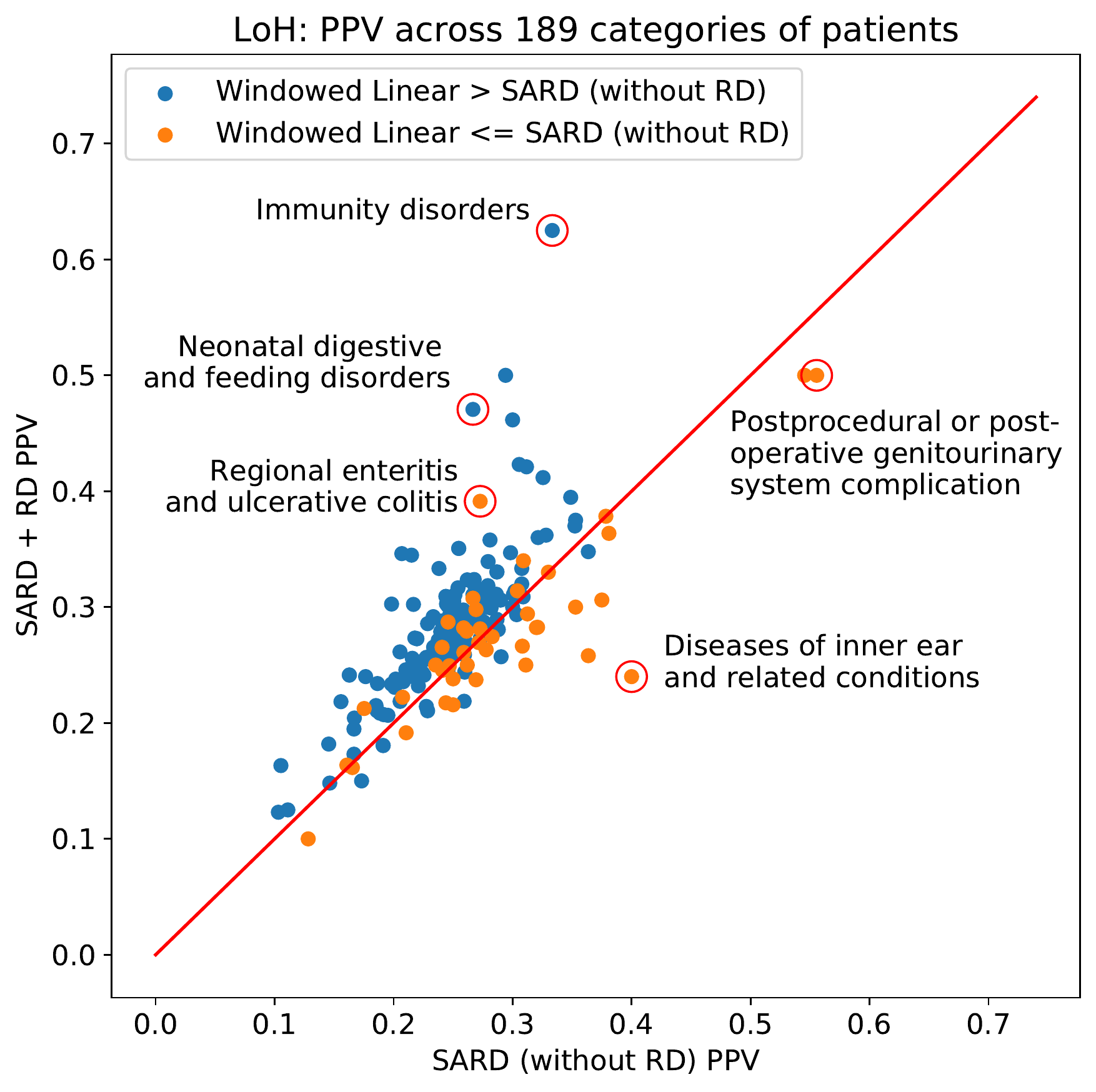}
  \caption{PPV for SARD with vs. without RD across subpopulations in the LoH task. Each point represents a patient category.}
  \label{figure:loh_subpopulation_ppv}
\end{figure}

{\bf Model performance across subpopulations.} For any clinical machine learning model, it is important to introspect on and be aware of differential performance across different groups of patients. We evaluate SARD's performance across a diverse range of patient clinical categories. We consider subpopulations defined by the Clinical Classifications Software Refined (CCSR) \cite{healthcare2020clinical} codes, and place a patient in a subpopulation if they experience at least three occurrences of a related condition within two years of prediction time. For the LoH task, Figure \ref{figure:loh_subpopulation_ppv} shows the positive predicted value (PPV, computed at a sensitivity of $0.5$ for each category and model) of SARD trained with and without RD across the 189 CCSR categories with at least 10 positive outcomes in the associated subpopulations. In addition to improvements in overall AUC, we find that SARD trained with RD outperforms a SARD model trained without RD in 147 out of 189 categories, spanning many diverse subpopulations, such as patients with immunity disorders and neonatal disorders.  

Figure \ref{figure:loh_subpopulation_ppv} also indicates whether the windowed linear baseline performed better than SARD without RD on each subpopulations, in terms of PPV. We find that for almost all categories where SARD outperforms SARD without RD, the linear baseline also outperforms. This corroborates our understanding that the success of SARD's unique pre-training procedure emanates from its ability to capture performant aspects of the linear baseline.

\subsection{Analyses of Reverse Distillation} \label{rd_experiment}
\begin{figure}
  \centering
  \includegraphics[width=.95\linewidth]{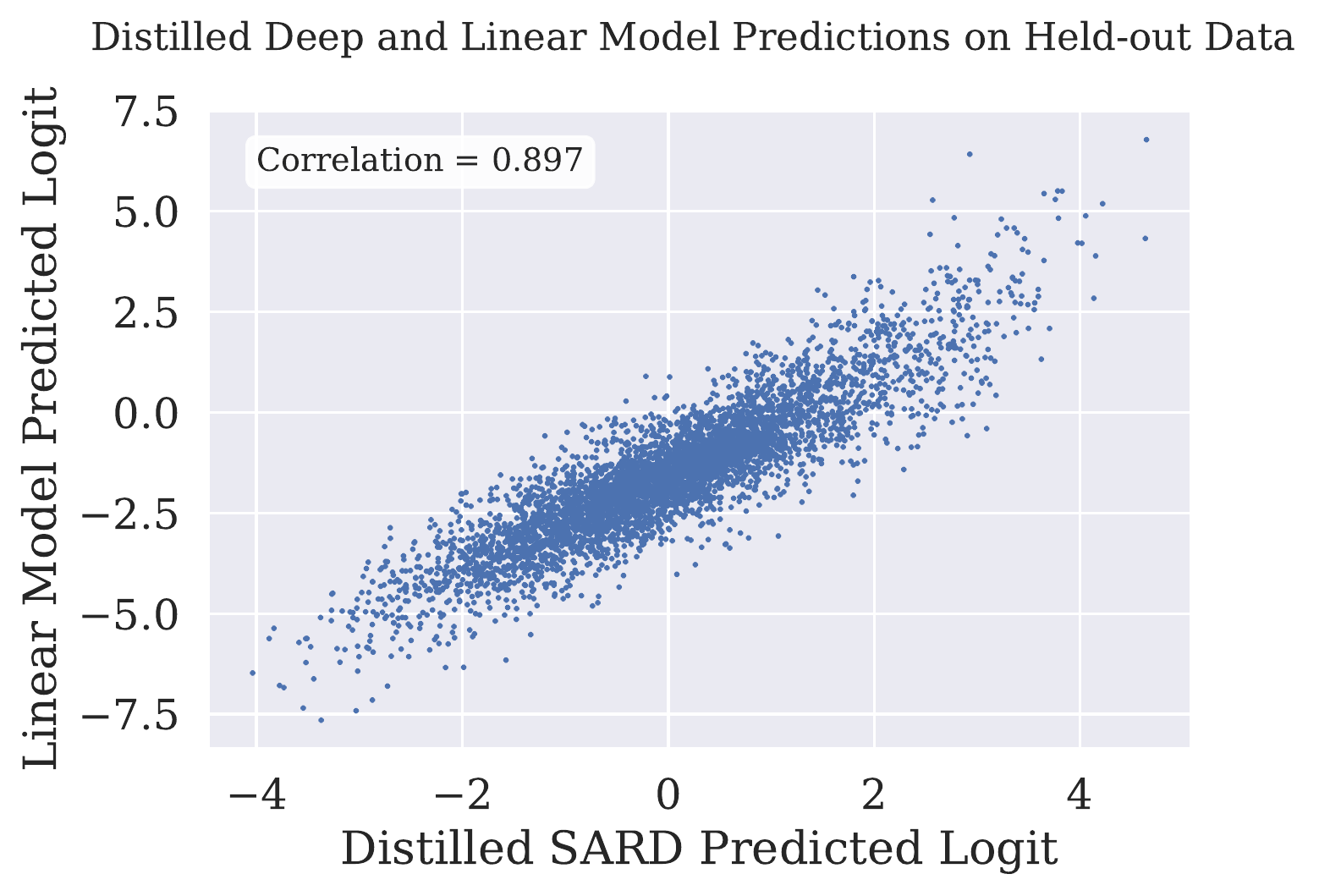}
  \caption{Comparison of Predictions on Held-out Data by Reverse Distilled and Linear Models}
  \label{figure:dist_lin_corr}
\end{figure}

We empirically validate that the SARD model for the End of Life task after reverse distillation (but before fine-tuning) generalizes in the same way as a linear model by analyzing the predictions made by both models on a held-out validation set. As seen in Figure \ref{figure:dist_lin_corr}, we find a Spearman correlation of $0.897$ between the logit outputs of the two models on held-out data\footnote{Recall that the logit corresponding to an output probability $p$ is $\log\left(p/(1-p)\right)$}. This indicates that even for unseen patients, the models make similar predictions. Thus, the reverse-distilled deep model does indeed mimic the linear model, not just memorize its outputs at certain points.

\begin{figure}
  \centering
  \includegraphics[width=.95\linewidth]{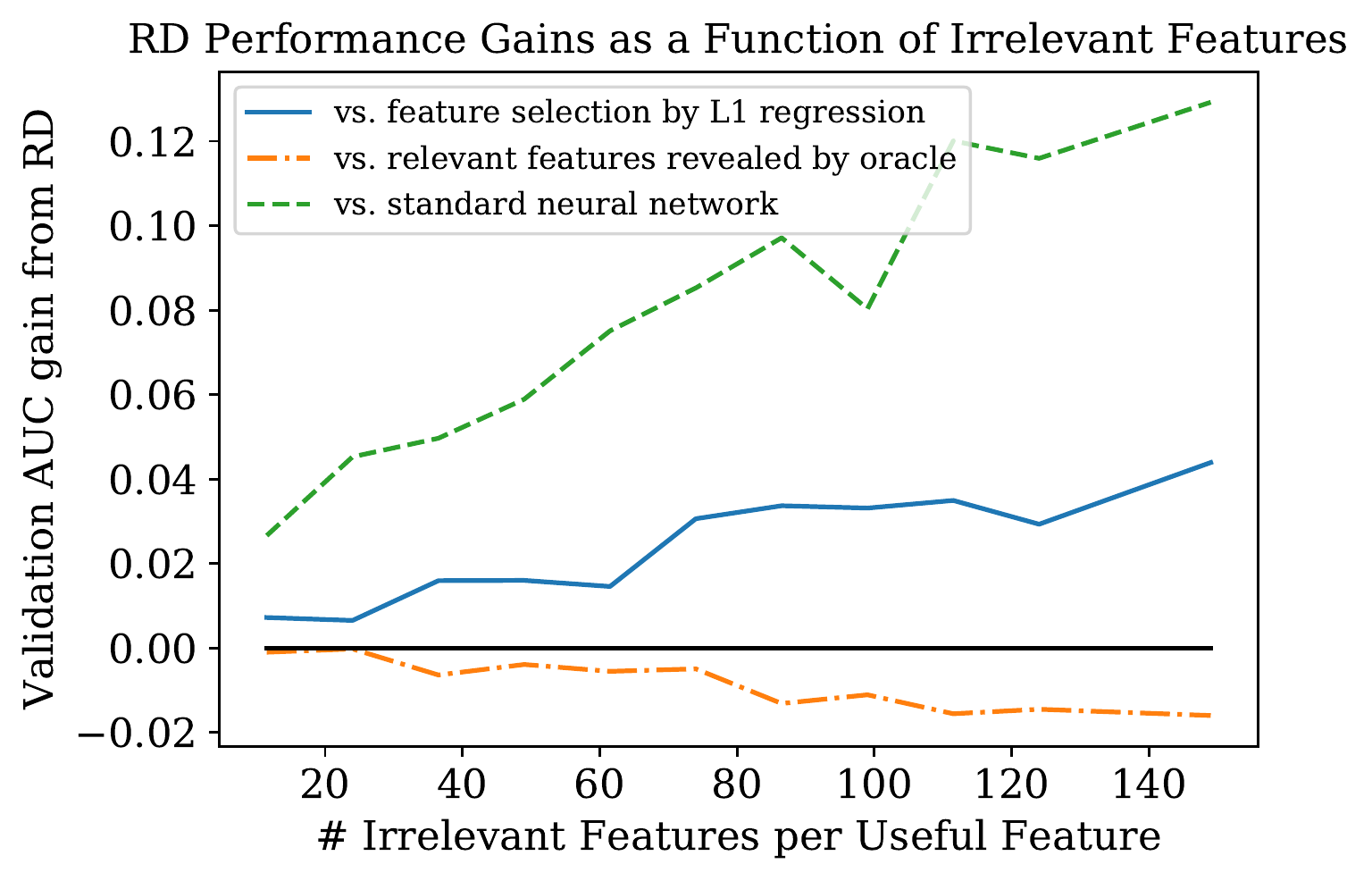}
  \caption{Reverse distillation AUC gains on synthetic data, as a function of sparsity of useful features}
  \label{figure:experiment_sparse_main}
\end{figure}

Reverse distillation is further analyzed via experiments on synthetic data in the Appendix. We find performance gains through reverse distillation for classification problems where data are poorly separated, or where only a small fraction of features are relevant, both properties of our prediction tasks. 
The ability of reverse distillation to enhance performance in synthetic scenarios with this property is shown in Figure \ref{figure:experiment_sparse_main}, where we additionally compare to alternative feature-selection methods. 

These experiments support that in addition to generalizing in the same way as an underlying linear model, a deep model trained via reverse distillation learns a soft version of the feature-selection performed by a regularized linear model. This is especially interesting in the case of multi-dimensional time-series data, where a simpler feature selection algorithm is not applicable. Indeed, in the case of longitudinal data, we would need to select a temporal context per feature, not just the features themselves. A naive approach of limiting SARD to the features selected by the windowed linear baseline in any time window results in no performance gains versus the baseline.



{\bf Network Dissection:} We present the results of our Network Dissection approach for intepretability. We summarize the findings of our correlation analysis as follows: for each neuron in SARD's penultimate layer, we ``match" it to the single linear model feature with which it had the highest MCC correlation; the linear model we refer to is the L1-regularized windowed logistic regression used for pre-training, and we only include features which have non-zero coefficients. In Table~\ref{table:uniqueconceptcorr} we report the total number of unique linear features which ``matched" at least one of the latent features in the penultimate layer of each deep model.

Unsurprisingly, we observe that the penultimate layers of our SARD networks trained \emph{without} RD pre-training do not capture a high fraction of the linear model's feature set. After RD pre-training, a much higher fraction of the linear model's features are represented by the penultimate layers of the deep models, and they remain so even after fine-tuning, highlighting RD's ability to effectively regularize even a fine-tuned model to make use of features known to be clinically meaningful. This helps explain the performance gains driven by reverse distillation seen in our experiments.

To better understand the impact of RD at the neuron level, we provide examples of top correlations for penultimate layer neurons trained with different SARD variants on the EoL task in the Appendix (see Tables \ref{table:corrA} and \ref{table:corrB}). 
For example, when training the network with RD, {\em before} fine-tuning, we find a neuron with correlation .487 with the linear model feature ``Hearing loss", .414 with ``Dementia", and .4 with ``Alzheimer's disease" (for all three, the $\infty$-time window). After fine-tuning, the same neuron has correlation .403 with ``Hearing loss", .32 with ``Dementia", and .312 with ``Subsequent hospital care", keeping the same broad interpretation although with a new emphasis on hospitalization. By contrast, none of the top 10,000 correlations for SARD trained {\em without} RD include a neuron correlated with the linear model feature for ``Hearing loss.”



\begin{table}[t]
\centering
\caption{Number (percentage) of unique linear model features represented by the final latent layer in the following model variants: SARD trained without RD pre-training (SARD (no RD)), SARD paused after pre-training (RD Only), and SARD with pre-training and fine-tuning (SARD).}
\label{table:uniqueconceptcorr}
\begin{tabular}{m{0.5\linewidth}m{0.1\linewidth}m{0.1\linewidth}m{0.1\linewidth}}\hline
\backslashbox{Model}{Task Name} & EoL & Surgery & LoH \\\hline\hline
 Total \# of Non-Zero Linear Features & 106  & 2000 & 1009\\\hline\hline
 SARD (no RD) & 41 (39$\%$)  & 52 (3$\%$) & 43 (4$\%$)\\\hline
 RD Only & 71 (67$\%$)  & 86 (4$\%$) & 161 (16$\%$)\\\hline
 SARD & 71 (67$\%$)  & 69 (3$\%$) & 144 (14$\%$)
 \\\hline
\end{tabular}
\end{table}




\section{Discussion}\label{conclusions}

We showed in Table \ref{table:auc} that two of the previous state-of-the-art deep models for longitudinal health data \citep{14,li2020behrt} do not outperform a well-tuned linear model with windowed features, consistent with previously reported results \citep[Supplemental Table 1]{2}. When trained without reverse distillation, our new architecture, SARD, achieves substantial wins in two of the tasks, yet also performs worse than the linear model on the third. However, when the models are pre-trained using reverse distillation, all of the architectures outperform the linear model, with SARD obtaining the best performance.
Reverse distillation is just one successful method by which self-attention based predictive models can be initialized. Although we did not observe an advantage in our dataset, possibly because of the small number of individuals relative to the large vocabulary, \citet{li2020behrt} demonstrated the use of masked language models as an unsupervised pre-training method for transformer-based models.

We hypothesize that reverse distillation will be of utility in other applications of deep learning with limited data where strong shallow models already exist. For example, within healthcare, interpretation of ECG waveforms (e.g. to predict atrial fibrillation) with deep models could be pre-trained with reverse distillation using linear models on easily derived clinical features such as R-R intervals \cite{Teijeiro_2018}. Beyond healthcare, text classification in under-resourced languages without pre-trained language models might benefit from reverse distillation using linear models with bag-of-words features.

We showed in Lemma \ref{lemma1} that our transformer architecture with temporal embeddings can represent a windowed linear model. However, that does not imply that gradient descent will learn a function that is equivalent to the linear model used within pre-training -- the objective is nonconvex and, even with infinite training data, there will be many equivalently good solutions. Nonetheless, we showed in Figure \ref{figure:dist_lin_corr} that the function learned by the deep model closely mirrors the function learned by the linear model on held-out data. A possible theoretical explanation might be found in recent work on convergence of stochastic gradient descent in over-parameterized deep models, coupled with the realization that pre-training is attempting to fit a particularly simple concept class, a linear model \cite{allenzhu2019convergence}.


\section*{Acknowledgements}
This work was supported by Independence Blue Cross and would not have been possible without the advice and support of Aaron Smith-McLallen, Ravi Chawla, Kyle Armstrong, Luogang Wei, and Jim Denyer. The Tesla K80s used for this research were donated by the NVIDIA Corporation.

\bibliography{references.bib}
\medskip

\clearpage
\begin{appendices}

\section{Choice of Baselines and Metrics}
As clinical prediction is a key task with numerous important applications, there exist numerous strong baselines for SARD to potentially be compared to. A natural comparison is BEHRT \cite{li2020behrt}, which recently outperformed previous state-of-the-art deep learning algorithms for medical records and served as the jumping-off point for our model; we describe our adaptation of BEHRT to our setting in detail in the Experiments section of our main paper. We also compare to RETAIN \cite{14,24}, as it is a deep learning model that, prior to BEHRT, achieved state-of-the-art performance on tasks similar to ours and offers an alternative way to use attention mechanisms to ingest longitudinal health data. Several other related methods were found to perform worse than BEHRT or RETAIN on longer term tasks, and as such we did not adapt them to our baselines. Other potential baselines use alternative types of EHR data, which were not available in our data -- for example, MIME \cite{choi2018mime} uses an EHR where treatments are explicitly justified with diagnoses. SARD instead uses less nuanced claims data that is representative of the input available in health insurance companies and large hospital systems.

We use the AUC-ROC score of our proposed methods and baselines on three predictive tasks as the primary metric of comparison. For completeness, we also provide a comparison of AUC-PRC scores in Table \ref{table:aucprc}. We find that SARD with RD continues to outperform baselines when using this alternative metric.

\begin{table}[h]
\centering
\caption{AUC-PRC Scores on Test Set}
\label{table:aucprc}

\begin{tabular}{m{0.5\linewidth}m{0.1\linewidth}m{0.1\linewidth}m{0.1\linewidth}}\hline
\backslashbox{Model}{Task Name}
&{EoL}&{Surgery}&{LoH}\\\hline\hline
$L_1$-reg. logistic regression \cite{4} & 0.099 & 0.834 & 0.202\\\hline \hline
RETAIN \cite{14} & 0.093 & 0.840 & 0.188\\\hline \hline
BEHRT \cite{li2020behrt} & 0.105 & 0.841 & 0.195 \\\hline
SARD (no RD) & 0.100 & 0.859 & 0.159 \\\hline
SARD (with RD) & \textbf{0.117} & \textbf{0.863} & \textbf{0.207} \\\hline

\end{tabular}
\end{table}





\section{Training Details and Hyperparameter Choices}\label{app:hyper}

We use a class weighting term $p_c$ equal to the ratio between the number of negative and positive training data points in both the pre-training and fine-tuning stages of SARD. Specifically, we pre-train our deep model on the loss function
\begin{eqnarray}
\ell_\textrm{RD}(x) &= &- p_cg_w(x) \log f_\theta(x) \nonumber \\ 
&&- (1-g_w(x))\log(1-f_\theta(x)),
\end{eqnarray}
and fine-tune the deep model on 
\begin{equation}
\ell_\textrm{tune}(x) = \ell_{CE}(x)+\alpha\ell_{RD}(x),
\end{equation}
where
\begin{eqnarray}
\ell_\textrm{CE}(x) &= &- p_cy(x) \log f_\theta(x) \\
    &&- (1-y(x)) \log(1-f_\theta(x)).
\end{eqnarray}
We display possible hyperparameter values for both our linear and SARD models in Table \ref{table:hyp}. These values were chosen using a validation set of $19,319$ patients. Here, $\mathcal{W}_c$ is the set of windows for our backwards-looking features, from which we chose a total of five windows, $\lambda$ is the inverse regularization constant for our $L_1$-regularized logistic regression baseline, and $\alpha$ is the weight placed on $\ell_\textrm{RD}$ during the fine-tuning stage of SARD. For our SARD models, we searched over attention depths $L$ of 1, 2 and 3; for our BEHRT baseline, we searched over $L$ of 2, 3, and 4.

As our models were trained on a small cluster of 4 GPUs, and our inputs were very high dimensional with around $1.9 \times 10^6$ features per person, it was critical for speed and feasibility to ingest data in a way that respected sparsity. As such, we made extensive use of \emph{scatter} operations to aggregate visits together. We note that this allows us to perform initial embeddings of concepts and times using a single large tensor in GPU memory, then summing up relevant terms for each visit. 

In addition, we accumulate gradients over multiple mini-batches to achieve a batch size of $500$. Indeed, the largest batch size we are able to operate with varied from 20 to 50 depending on the model in question. 

\begin{table}[h]
\centering
\caption{Hyperparameter values searched during tuning.}
\label{table:hyp}
\begin{tabular}{m{0.26\linewidth}|m{0.7\linewidth}}\hline
{Hyperparameter}&{Possible Values} \\\hline\hline
$\mathcal{W}_c$ & $[T_A-t',T_A]$ for all $t' \in \{15, 30, 60, 90, 180, 360, 540, 720, \infty\}$ days, where $T_A$ represents the prediction date \\\hline
$\lambda$ & $\{20, 2, 0.2, 0.02, 0.002, 0.0002\}$ \\\hline
$\alpha$ & $\{0, 0.05, 0.1, 0.15, 0.20$\}\\\hline
\end{tabular}
\end{table}

\section{Proof Sketch of Lemma 1} \label{app:lemmaproof}

We show that a single self-attention head can generate the vector $f_i(W)$, of size $|f_i(W)|$, for a given window $[T_A-T,T_A]$ as described in our main paper's `Learning with Reverse Distillation' section, thus implying that several self-attention heads' concatenated outputs can generate the concatenation of several $f_i(W)$ vectors.

Set the embedding function $\phi(c)$ to simply return a one-hot encoding of the code $c$, concatenated with $d_e$ zeros. We further set the linear model $g$ to be an identity function. Then for all $i,j$, the embedded visit content vector $\psi(V^i_j)$ will be a multi-hot binary vector whose nonzero elements correspond to the codes in $C^i_j$.

We note that our time embedding per visit will be  $\tau\left(V^i_j\right)=sin\left(t^i_j\mathbf{\omega}\right)||cos\left(t^i_j \mathbf{\omega}\right)$. We set the first $|\mathcal{C}|$ elements of $\omega$ to zero, so that visit embedding $\psi(V^i_j) + \tau\left(V^i_j\right)$ will be fully separable component-wise into a multi-hot vector of codes and a time embedding.

The self-attention mechanism will use a linear map from $\psi(V^i_j) + \tau\left(V^i_j\right)$ to three vectors $k^i_j,q^i_j,v^i_j$ called the key, query and value vectors respectively, and create the contextual embedding $\sum_{j'=1}^{n_v} \left(q^i_j\cdot k^i_{j'}\right)v^i_{j'}$ for visit $V^i_j$. We allow $v^i_j$ to simply be the multi-hot encoding $\psi(V^i_j)$. Note that since $\psi(V^i_j)$ and $\tau(V^i_j)$ have different nonzero components that this can be achieved by a simple matrix multiplication from $\psi(V^i_j) + \tau\left(V^i_j\right)$.

Next, we create appropriate length-$1$ key and query vectors. We define $k^i_j=q^i_j = [[t^i_j < T]]$, and under this definition the contextual embedding of every visit $V^i_j$ where  $t^i_j < T$ will become $\sum_{j'=1}^{n_v} [[t^i_{j'}<T]] v^i_{j'}$, which is a multi-hot vector whose nonzero entities correspond to all codes seen in the window of the past $T$ days.

It remains to show how we would construct $k^i_j=q^i_j = [[t^i_j < T]]$ as a linear transformation of $\psi(V^i_j) + \tau\left(V^i_j\right)$. We do so by invoking a Fourier analysis argument. Let $P$ be the length of the interval from the first event in the dataset to $T_A$. Then, $[[t^i_j < T]]$ can simply be represented as a function of period $P$ with value $1$ in $[0,T]$ and $0$ in $[T,P]$, which in turn can be represented as a Fourier series with coefficient $\frac{2}{n\pi}\sin^2(\frac{n \pi T}{P})$ corresponding to $\sin(\frac{2 n \pi t}{P})$ and coefficient $\frac{1}{n\pi}\sin(\frac{n \pi T}{P})$ corresponding to $\cos(\frac{2 n \pi t}{P})$, Thus, for appropriately chosen $\omega$ that includes values of the form $2 n \pi / P$, we can recover an arbitrarily good approximation of $[[t^i_j < T]]$, thus allowing us to use a single self-attention head to mimic a single windowed feature vector as passed into the linear model. The convolutional prediction head can simply apply an identity transformation by setting each of of $|f_i(W)|$ kernels equal to the vectors of the standard basis for $\mathbb{R}^{|f_i(W)|}$. Noting that the softmax function on the attention weights is largely irrelevant if we scale up the pre-softmax representation of each visit, the nonzero terms will dominate even after the application of softmax function, and therefore after the max pooling operation filters corresponding to nonzero elements of $f_i(W)$ will return values close to one, and those corresponding to elements of $f_i(W)$ will return values close to $0$, effectively implementing the identity transformation on this binary vector. 

Using multiple self-attention heads, we can obtain the concatenation of several windowed feature vectors, and passing these through the prediction head allows us to fully replicate the functionality of the linear model using the deep model.

\section{Model Transfer Across Time}\label{app:shift}
Healthare data is generally considered to be non-stationary, in that the distribution of codes observed in EHR data will shift over time due to shifts in populations, technologies, and best practices \cite{jung2015implications}. With this in mind, we evaluate how well our SARD model trained on data from one time period transfers to the future. We train our models for all tasks on data from 2016 and earlier and have a prediction date of January 1, 2017, on which it predicted deaths occuring between April and September of 2017. The same model is then used to predict outcomes at various times in the future, as seen in Table \ref{table:auc_nonstat}. We find that the model is still able to outperform linear baselines. We believe this is partly due to the well-regularized nature of our models. 


\begin{table}[t]
\centering
\caption{AUC-ROC Scores for EoL on Test Sets using Jan 1, 2017 Model}
\label{table:auc_nonstat}
\begin{tabular}{m{0.25\linewidth}m{0.15\linewidth}m{0.15\linewidth}m{0.20\linewidth}}\\\hline 
{Prediction Date}&{SARD}&{SARD (no RD)}&{Windowed Linear Baseline}\\\hline\hline
Jan 1, 2017 & 85.6 & 85.0 & 83.4\\\hline
July 1, 2017 & 83.9 & 83.3 & 82.0 \\\hline
Jan 1, 2018 & 84.4 & 83.8 & 82.4 \\\hline
July 1, 2018 & 84.5 & 83.9 & 82.8 \\\hline
\end{tabular}
\end{table}

\section{SARD Extracted Timeline Case Study}\label{app:case}
In this appendix we qualitatively investigate the ability of SARD to generalize and contextualize better than a linear model through a case study.

In order to convert the soft predictions of SARD and our baseline windowed linear model to binarized predictions of outcomes, we chose decision thresholds for both models to ensure a false positive rate of $0.25$ on a validation set -- this resulted in thresholds of $0.34$ and $0.33$ for SARD and the linear model respectively. We note that in practice the selection of a threshold, or alternatively the use of these scores as rankings, would be driven by varied downstream applications. 

We consider a female patient who died between April and September of 2017 (she was $\geq$90 years old at the time), an event that was correctly predicted by SARD (probability of $71.1\%$) but not by our linear baseline (probability of $5.4\%$). She had an active medical history, with over 700 recorded medical visits. To better understand why SARD accurately predicts her death while logistic regression does not, we introspect on which visits are most influential in the prediction head of the model and how those visits were benefited by the self-attention architecture.

To determine which visits were most influential in the prediction made by SARD, we use the score $s(V_j) = \sum_k [[V_j=\nu_k]] w_k \sigma(\chi_k)$ to assign an importance to visit $V_j$, as defined in the `Model Introspection' section of our main paper.

The specific visit that maximizes the above score, which we denote as the \emph{top visit}, occurred in 2016, a few months before the patient's death. During this visit, she was treated for a \texttt{chronic ulcer of skin of lower leg}, with both \texttt{anesthesia procedure(s)} and \texttt{Debridement, muscle and/or fascia} procedures performed. The visit also notes that the patient experienced  \texttt{atrial fibrillation}, \texttt{Pure hypercholesterolemia}, \texttt{Non-rheumatic aortic sclerosis}, and \texttt{ hypothyroidism}, and that an \texttt{old myocardial infarction} was detected. Further details of this visit are provided in Table \ref{table:topshapvisits}.

While some of the codes in this visit, such as the treatment of an ulcer, are not immediately alarming in their own right and are therefore not highly weighted by the linear model, this patient's history of ongoing cardiovascular disease suggest that these conditions may indeed be manifestations of more serious underlying pathology. Aligning with medical intuition that the long-term survival rate of patients who suffer a myocardial infarction is highly dependent upon other risk factors \cite{martinlong}, the linear model's top weighted negative features are \texttt{Insertion and placement of flow directed catheter (Swan-Ganz)}, a procedure used diagnostically to determine and eliminate risks after a myocardial infarction, and \texttt{carvedilol}, a drug known to reduce risk of death after myocardial infarction, both over length-$\infty$ windows. 
  
\begin{figure}
    \centering
    \includegraphics[scale=0.35]{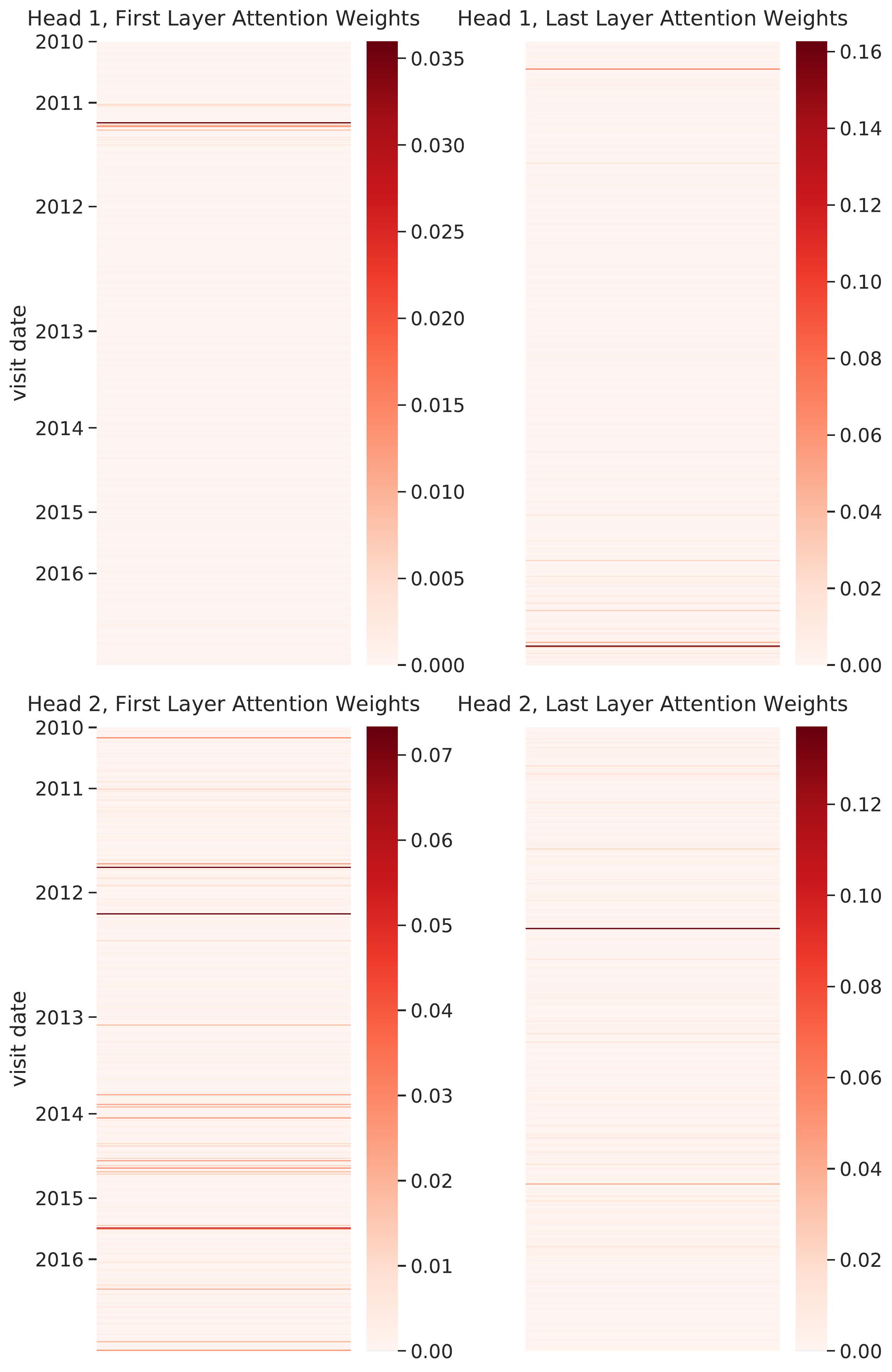}
    \caption{Attention weights for the case study patient's `top visit.' While the top visit occurred in 2016, it pulls context from visits throughout the patient's history. Each panel contains a row for each of the patient's 512 visits, colored by how much attention it is given by the top visit. Notably, the visits most highly attended to in the first layer of the first self-attention head (top left panel) represent a serious prior manifestation of the same underlying atherosclerotic vascular disease present in the top visit.}
    \label{fig:case_study_all4heads}
\end{figure}

The SARD model is able to leverage important contextual information from throughout the patient's history thanks to its self-attention mechanism. While the top visit occurred in 2016, SARD is able to build an understanding that the patient was at high cardiovascular risk at the time of prediction by strongly attending to visits encoding severe cardiovascular illness in 2011, as visualized in Figure \ref{fig:case_study_all4heads}. Indeed, all of the visits strongly attended to by the first layer of the first self-attention head of SARD for this patient are related to this 2011 episode and involve major cardiovascular events, as seen in Table \ref{table:topattnvisits}. These persistent manifestations of the patient's underlying cardiovascular disease provide context for more recent visits, and augment the threat posed by her recent cardiovascular disease at prediction time.

\begin{table*}[]
\small
\caption{The 4 most predictive visits for the case study patient, ranked by the score $s(V_j)$ developed in main paper section `Model Introspection.' While our models use specific visit dates, we only include visit year to censor protected health information (PHI).}
\label{table:topshapvisits}
\begin{tabular}[c]{|p{0.95\linewidth}|}
\hline
\small
  
  \begin{tabular}[c]{p{0.95\linewidth}}
  
  \textbf{Visit Importance Score}: 0.487 \\
  \textbf{Year of Visit}: 2016 \\
  \textbf{Codes:}  Hypothyroidism | 
 Calcium Chloride 0.0014 MEQ/ML / Potassium Chloride 0.004 MEQ/ML / Sodium Chloride 0.103 MEQ/ML / Sodium Lactate 0.028 MEQ/ML Injectable Solution | 
 Anesthesia for procedures on the integumentary system on the extremities, anterior trunk and perineum; not otherwise specified | 
 Unlisted anesthesia procedure(s) | 
 Debridement, muscle and/or fascia (includes epidermis, dermis, and subcutaneous tissue, if performed); first 20 sq cm or less | 
 Atrial fibrillation | 
 Old myocardial infarction | 
 2 ML Fentanyl 0.05 MG/ML Injection | 
 General Surgery | 
 Anesthesiology | 
 20 ML Propofol 10 MG/ML Injection | 
 Debridement, muscle and/or fascia (includes epidermis, dermis, and subcutaneous tissue, if performed); each additional 20 sq cm, or part thereof (List separately in addition to code for primary procedure) | 
 Non-rheumatic aortic sclerosis | 
 Chronic ulcer of foot | 
 Pure hypercholesterolemia | 
 Necrosis of ankle muscle co-occurrent and due to chronic ulcer of ankle | 
 Chronic ulcer of skin of lower leg | 
 Cefazolin 1000 MG Injection | 
 Osteoporosis
  \end{tabular} 
  \\ \hline

  \begin{tabular}[c]{p{0.95\linewidth}}
  
  \textbf{Visit Importance Score}: 0.467 \\
  \textbf{Year of Visit}: 2015 \\
  \textbf{Codes:}  Hypothyroidism | 
 Collection of venous blood by venipuncture | 
 Immunization administration (includes percutaneous, intradermal, subcutaneous, or intramuscular injections); 1 vaccine (single or combination vaccine/toxoid) | 
 Office or other outpatient visit for the evaluation and management of an established patient, which requires at least 2 of these 3 key components: A detailed history; A detailed examination; Medical decision making of moderate complexity. Counseling and/o | 
 Essential hypertension | 
 Cardiology | 
 Clinical Laboratory | 
 0.5 ML influenza B virus vaccine B/Brisbane/60/2008 antigen 0.12 MG/ML / Influenza Virus Vaccine, Inactivated A-California-07-2009 X-179A (H1N1) strain 0.12 MG/ML / Influenza Virus Vaccine, Inactivated A-Victoria-210-2009 X-187 (H3N2) (A-Perth-16-2009) st | 
 Paroxysmal atrial fibrillation | 
 Iron deficiency anemia | 
 Pure hypercholesterolemia
  \end{tabular} 
  \\ \hline

  \begin{tabular}[c]{p{0.95\linewidth}}
  
  \textbf{Visit Importance Score}: 0.345 \\
  \textbf{Year of Visit}: 2016 \\
  \textbf{Codes:}  
   10 ML Nitroglycerin 5 MG/ML Injection | 
 Ulcer of lower extremity | 
 Anesthesia for diagnostic arteriography/venography | 
 Unlisted anesthesia procedure(s) | 
 Aortography, abdominal, by serialography, radiological supervision and interpretation | 
 Angiography, extremity, bilateral, radiological supervision and interpretation | 
 Anesthesia for patient of extreme age, younger than 1 year and older than 70 (List separately in addition to code for primary anesthesia procedure) | 
 5 ML Fentanyl 0.05 MG/ML Injection | 
 Anesthesiology | 
 Vascular Surgery | 
 Lidocaine Hydrochloride 10 MG/ML Injectable Solution | 
 Revascularization, endovascular, open or percutaneous, femoral, popliteal artery(s), unilateral; with transluminal angioplasty | 
 Chronic atrial fibrillation | 
 heparin sodium, porcine 1000 UNT/ML Injectable Solution | 
 Atherosclerosis of native arteries of the extremities | 
 Cefazolin 1000 MG Injection
  \end{tabular} 
  \\ \hline
  
   \begin{tabular}[c]{p{0.95\linewidth}}
  
  \textbf{Visit Importance Score}: 0.337 \\
  \textbf{Year of Visit}: 2011 \\
  \textbf{Codes:}   Acquired hypothyroidism | 
 Neurogenic bladder | 
 Peptic ulcer without hemorrhage, without perforation AND without obstruction | 
 Radiologic examination, chest; single view, frontal | 
 Initial hospital care, per day, for the evaluation and management of a patient, which requires these 3 key components: A comprehensive history; A comprehensive examination; and Medical decision making of high complexity ...| 
 Inpatient consultation for a new or established patient, which requires these 3 key components: A comprehensive history; A comprehensive examination; and Medical decision making of high complexity ... | 
 Emergency department visit for the evaluation and management of a patient, which requires these 3 key components within the constraints imposed by the urgency of the patient's clinical condition and/or mental status: ... | 
 Dyspnea | 
 Atrial fibrillation | 
 Aortic valve disorder | 
 Heart murmur | 
 Low blood pressure | 
 Hypertensive heart disease without congestive heart failure | 
 Coronary atherosclerosis | 
 Disorder of kidney and/or ureter | 
 Ankle ulcer | 
 Coronary arteriosclerosis in native artery | 
 Dizziness and giddiness | 
 Acute myocardial infarction of anterior wall | 
 Leukocytosis | 
 Tachycardia | 
 Chest pain | 
 Osteoporosis
  \end{tabular} 
  \\ \hline

\end{tabular}
\end{table*}

\begin{table}[]
\small
\caption{Three visits most highly attended by the top visit from 2016; first layer, first self-attention head. We only include visit year in the table so as not to reveal PHI.}
\label{table:topattnvisits}
\begin{tabular}[c]{|p{0.95\linewidth}|}
\hline
\small
  
  \begin{tabular}[c]{p{0.95\linewidth}}
  
  \textbf{Attention from Top Visit}: 0.0360 \\
  \textbf{Year of Visit}: 2011 \\
  \textbf{Codes:}  Disorder of cardiovascular system | 
 Echocardiography, transthoracic, real-time with image documentation (2D), includes M-mode recording, when performed, complete, with spectral Doppler echocardiography, and with color flow Doppler echocardiography | 
 Duplex scan of extracranial arteries; complete bilateral study | 
 Subsequent hospital care, per day, for the evaluation and management of a patient, which requires at least 2 of these 3 key components: An expanded problem focused interval history; An expanded problem focused examination ... | 
 Acute myocardial infarction | 
 Angina pectoris
  \end{tabular} 
  \\ \hline
  \begin{tabular}[c]{p{0.95\linewidth}}
  
  \textbf{Attention from Top Visit}: 0.0130 \\
  \textbf{Year of Visit}: 2011 \\
  \textbf{Codes:}  Radiologic examination, chest; single view, frontal | 
 Atelectasis | 
 Aortic valve disorder
  \end{tabular}
  \\ \hline
  \begin{tabular}[c]{p{0.95\linewidth}}
  
  \textbf{Attention from Top Visit}: 0.0086 \\
  \textbf{Year of Visit}: 2011 \\
  \textbf{Codes:} Open and other replacement of aortic valve with tissue graft | 
 Anesthesia for direct coronary artery bypass grafting; with pump oxygenator | 
 Replacement, aortic valve, open, with cardiopulmonary bypass; with prosthetic valve other than homograft or stentless valve | 
 Arterial catheterization or cannulation for sampling, monitoring or transfusion (separate procedure); percutaneous | 
 Radiologic examination, chest; single view, frontal | 
 Level IV - Surgical pathology, gross and microscopic examination Abortion - spontaneous/missed Artery, biopsy Bone marrow, biopsy Bone exostosis Brain/meninges, other than for tumor resection Breast, biopsy, not requiring microscopic evaluation of surgica | 
 Decalcification procedure (List separately in addition to code for surgical pathology examination) | 
 Insertion and placement of flow directed catheter (eg, Swan-Ganz) for monitoring purposes | 
 Anesthesia for patient of extreme age, younger than 1 year and older than 70 (List separately in addition to code for primary anesthesia procedure) | 
 Atrial fibrillation | 
 Aortic valve disorder | 
 Essential hypertension | 
 Coronary atherosclerosis | 
 Operative external blood circulation | 
 Packed blood cell transfusion | 
 Echocardiography | 
 Chest pain
  \end{tabular}
  \\ \hline

\end{tabular}
\end{table}

\section{Reverse Distillation on Synthetic Data}\label{app:syn}
It is not immediately obvious why reverse distillation helps model performance when utilized as a pre-training procedure. To help further empirically justify when and how reverse distillation works, we turn to experiments with synthetic data designed to mimic the distinct properties of the kind of data found in electronic health records. In particular, we are interested in data where:
\begin{itemize}
    \item The data is high-dimensional but only a small fraction of these features are useful for any specific downstream task. 
    \item The data is not fully separable, even in the limit of infinite data.
\end{itemize}
With these two properties in mind, synthetic data for a binary classification problem is generated as follows:
\begin{itemize}
    \item First, two centers $c_0,c_1$ are chosen in $\mathbb{R}^d$, with a separation of $||c_0-c_1|| = \gamma$. We shift the clusters so that the origin is exactly between the two centers.
    \item Next, for each of N training points:
        \begin{itemize}
            \item We draw a label $y$ for the point from a Bernoulli distribution with parameter $\rho$.
            \item We associate $K$ features with the point. The first $\beta K$ are drawn as iid Gaussian RVs with mean $c_y$ and unit variance. The remainder are uninformative features drawn as iid Gaussians with mean 0 and unit variance.
        \end{itemize}
\end{itemize}

We can manipulate the fraction of useful features and the separability of the classes by varying $\beta$ and $\gamma$ respectively. Our experiments are designed to find when reverse distillation is successful in excess of a simple feature selection procedure -- the hypothesis is that reverse distillation would put weight on features similar to those chosen by the underlying linear model, but in a `soft' and more robust way. As such, the baseline we choose to compare to is a deep model trained only on the features that are not zeroed out by a  $L_1$-regularized linear model.

We note that other, more complex feature selection baselines are possibilities. However, feature-selection in general is straightforward to implement in this synthetic model – one can simply slice out the features chosen by a procedure. With longitudinal medical data, we are not just selecting features but temporal contexts as well, and it is not possible to iterate over all such selections. As such, reverse distillation lets us do a “soft” feature selection over a very complex space of time-series features. 

Concretely, we define four procedures whose performance we compare:
\begin{itemize}
    \item Reverse Distill: We first train an $L_1$-regularized logistic regression on a synthetic binary classification dataset, tuning the regularization with a validation set to maximize AUC. We collect the predictions $p_{LR}(x)$ made by the linear model at each training point $x$. Next, a multi-layer perceptron (MLP) with two densely connected layers with ReLU activation, followed by a sum and sigmoid activation to return a probability is initialized randomly and trained until convergence to minimize the KL-divergence between its predictions $MLP(x)$ and $p_{LR}(x)$. Finally, this MLP is fine-tuned by minimizing the loss $$\ell(x,y)= xent(MLP(x),y)+ \alpha D_{KL}(p_{LR}(x)|MLP(x))$$
    where $\alpha$ is a hyperparameter tuned on a validation set.
    \item Standard Neural Network: We create an MLP using the same architecture as Reverse Distill, and train it until convergence to minimize the loss $\ell(x,y)= xent(MLP(x),y)$.
    \item Feature Selection by $L_1$ regression: We first train an $L_1$-regularized logistic regression as in Reverse Distill. Denote the weights of this model by $w_i$ -- we define a feature selection function $f_R(x) = \left< x_i \right>_{\{i|w_i\neq 0\}}$ which takes a feature vector $x$ and creates a new vector whose components correspond to the elements of $x$ which would not be zeroes out by the regularized logistic regression. We create an MLP using the same architecture as Reverse Distill, and train it until convergence to minimize the loss $\ell(x,y)= xent(MLP(f_R(x),y)$
    \item Feature Selection by Oracle: We define a feature selection function $f_O(x) = \left< x_i \right>_{\{i|\texttt{feature i is relevant}\}}$ which takes a feature vector $x$ and creates a new vector whose components correspond to the $\beta K$ elements of $x$ which are actually relevant for prediction. We create an MLP using the same architecture as Reverse Distill, and train it until convergence to minimize the loss $\ell(x,y)= xent(MLP(f_O(x),y)$. This model reflects an optimal feature selection procedure only possible with full knowledge of the generative process for the data, and should beat all other baselines. 
\end{itemize}
We compare the differences in median AUCs on out-of-sample data between Reverse Distill and the other three models to investigate when reverse distillation is useful. Unless explicitly varied, we hold the data generation parameters at $K=200,\gamma=0.5,\rho=0.05,\beta=0.02$:
\begin{itemize}
    \item We first investigate how the separability of the data affects the performance gains of reverse distillation by varying $\gamma$. We expect that at extremely low separability, no model will be able to do well, and at high separability all models will do equally well. Between these two extremes, we expect reverse distillation to outperform baselines. This is confirmed by our experimental results, as visualized in Figure  \ref{figure:experiment_sep}
    
    \begin{figure}[h]
      \centering
      \includegraphics[width=0.99\linewidth]{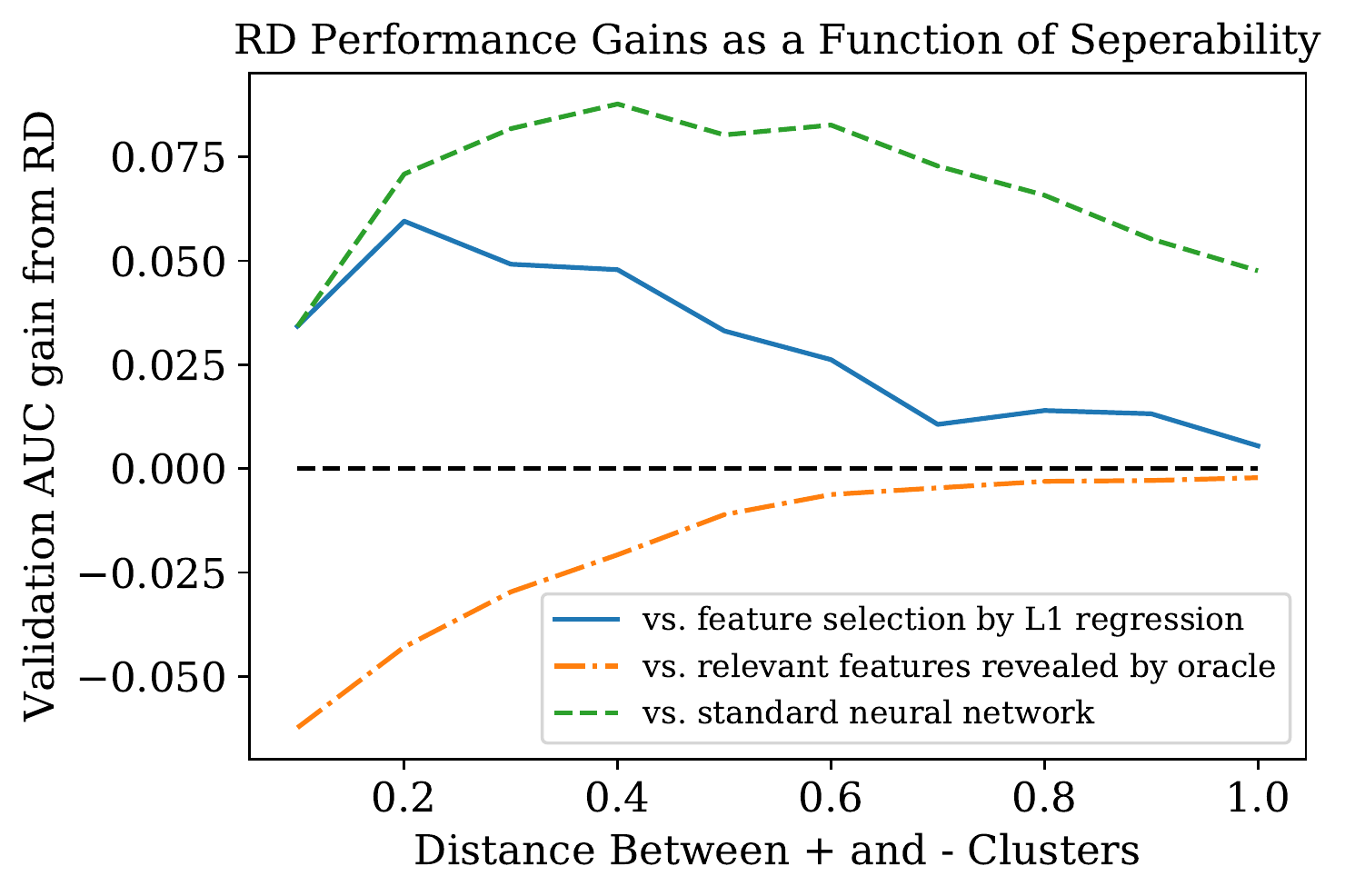}
      \caption{Reverse Distillation performance gains as a function of class separability}
      \label{figure:experiment_sep}
    \end{figure}
    
    \item We next investigate how the sparsity of useful features affects the performance gains of reverse distillation by varying $\alpha$. We expect that as $\alpha$ decreases and we see less useful features, that reverse distillation will be more useful since it can make nuanced soft feature selections that can greatly help downstream performance. This is confirmed by our experimental results, as visualized in Figure  \ref{figure:experiment_sparse}
    
    \begin{figure}[h]
      \centering
      \includegraphics[width=0.99\linewidth]{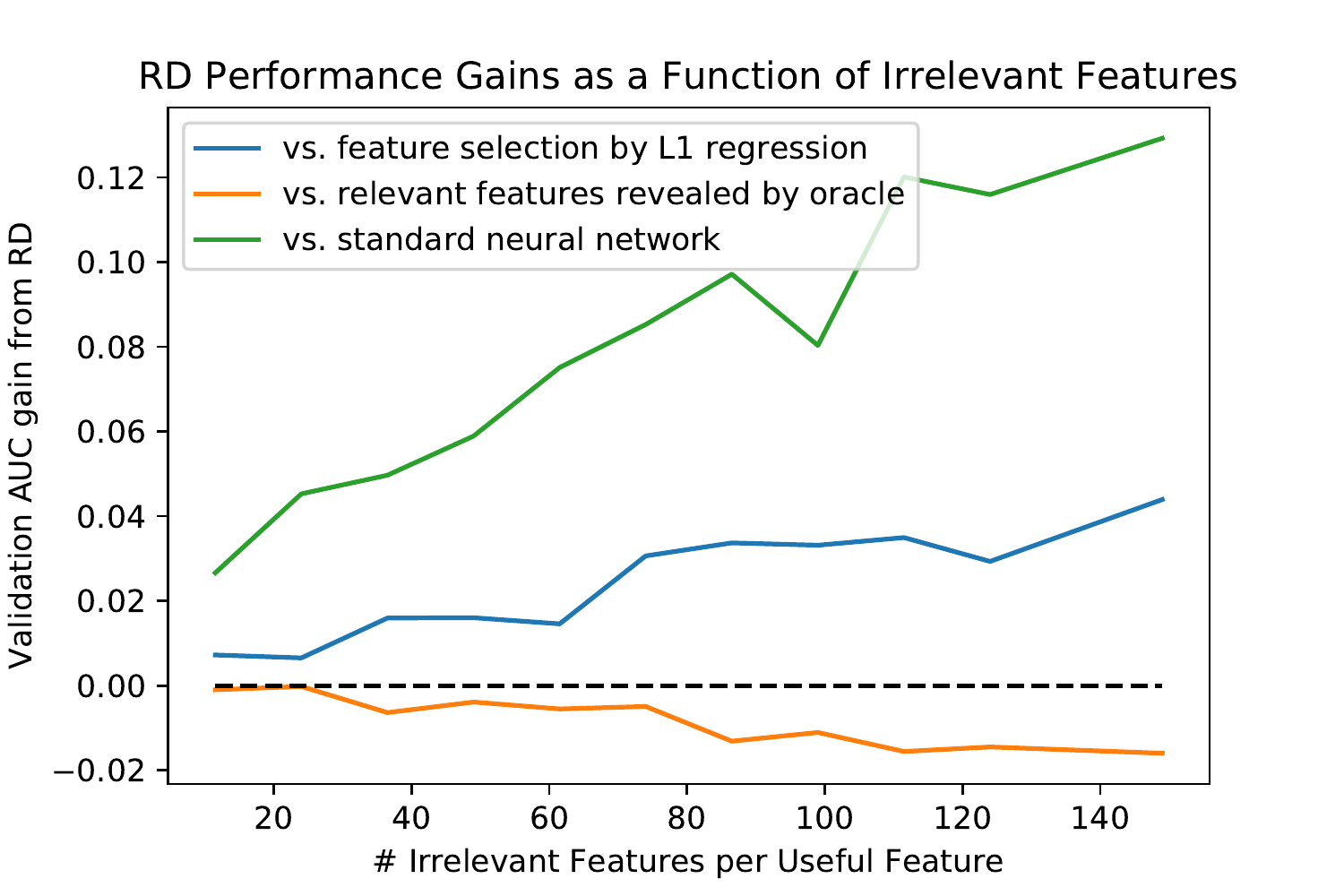}
      \caption{Reverse Distillation performance gains as a function of sparsity of useful features}
      \label{figure:experiment_sparse}
    \end{figure}
    
    \item We find that reverse distillation performs better in the regime where many features are redundant, as measured by varying $\beta$. We expect that reverse distillation can effectively isolate relevant features by learning to use the same features isolated by the $L_1$ regression. This is verified by our experimental results shown in Figure  \ref{figure:experiment_sparse_main}

    \begin{figure}[h]
      \centering
      \includegraphics[width=0.99\linewidth]{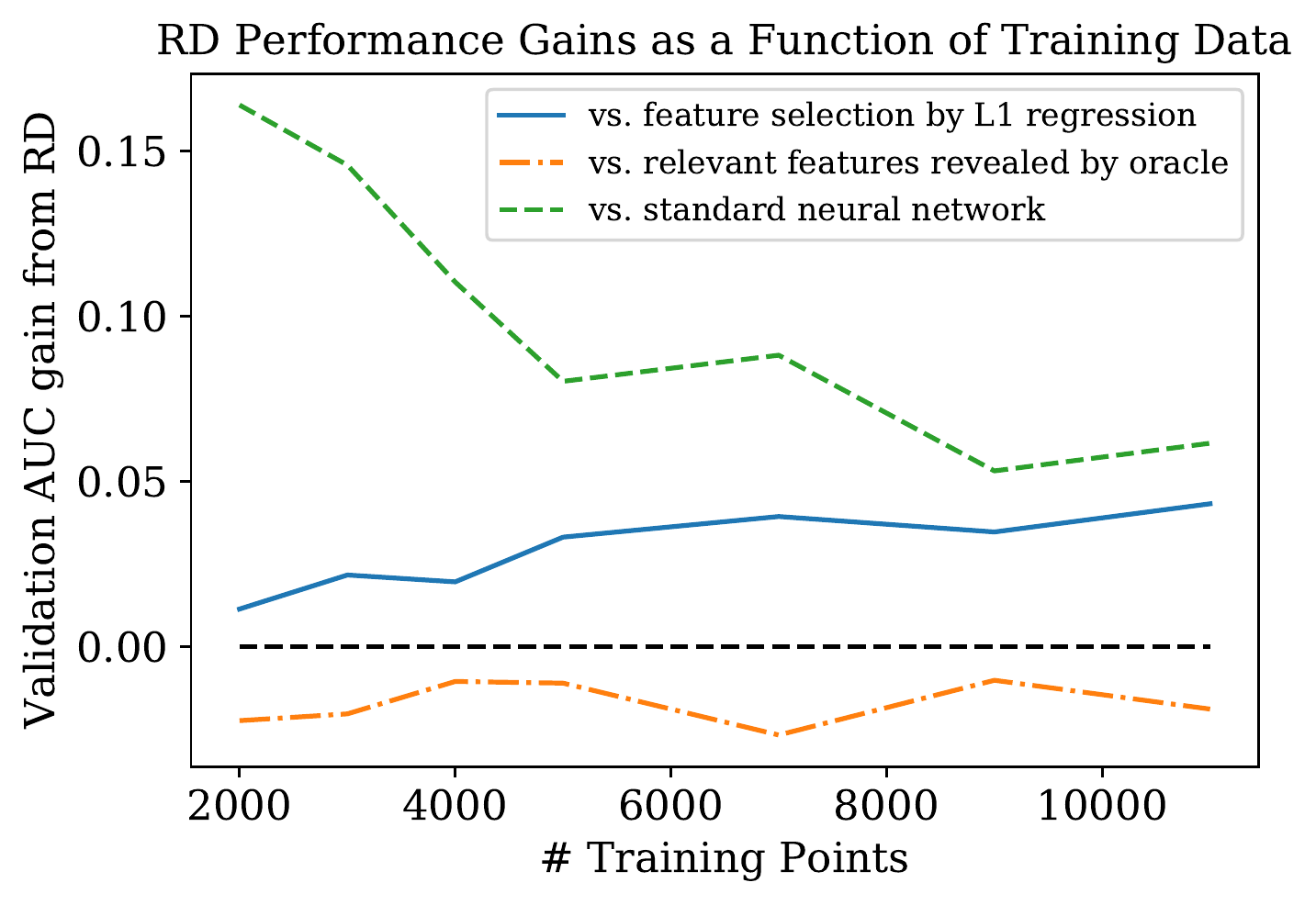}
      \caption{Reverse Distillation performance gains as a function of amount of training data}
      \label{figure:experiment_samples}
    \end{figure}
\end{itemize}

\section{Additional Data from Network Dissection} \label{app:dissect}

As described in our main paper's `Interpretability via Network Dissection' and `Network Dissection' sections, we analyze the penultimate layer neurons from several SARD network variants trained on the EoL task in terms of their correlations with the features of the linear model. Specifically, we can examine how the correlations on the \textit{same} neuron change between pre-training outputs (referred to as RD Only) and after fine-tuning (referred to as SARD, as this corresponds to SARD's general training procedure). Table \ref{table:corrA} shows the top five correlated features for the neuron whose most correlated linear model feature both immediately after RD pre-training and after fine-tuning is ``hearing loss". By contrast, none of the top 10,000 correlations for SARD without RD (referred to as SARD (no RD)) involve the feature ``hearing loss". 

As another example, Table \ref{table:corrB} shows the neuron most correlated with ``coronary atherosclerosis" in each SARD network. We see that the RD Only and SARD models have neurons that correlate with the same top 5 linear features, which are all highly specifically related to coronary atherosclerosis, for example ``coronary arteriosclerosis in native artery" and ``mitral valve disorder." By contrast, the SARD (no RD) neuron which is most correlated with the linear feature ``coronary atherosclerosis" is also highly correlated with other non-specific features, such as ``subsequent hospital care," ``initial hospital care" and ``hospital discharge day management." Without RD, the deep model did not successfully learn a neuron corresponding to this specific and highly weighted linear model feature.



\begin{table*}[ht]
\centering
\caption{Top 5 Correlated features for an example neuron from the penultimate layer of SARD models trained on the EoL task. RD Only refers to the model immediately after the pre-training stage, and SARD refers to the model after fine-tuning.}
\label{table:corrA}
\begin{tabular}{m{0.2\linewidth}m{0.6\linewidth}m{0.1\linewidth}}\hline
Model & Feature Name & Correlation \\\hline\hline
 & Hearing loss (10000 days) & 0.487  \\
& Dementia (10000 days) & 0.414 \\
RD Only & Alzheimer's disease (10000 days) & 0.400 \\
& Altered mental status (10000 days) & 0.308 \\
& Muscle weakness (10000 days) & 0.302 
\\\hline
& Hearing loss (10000 days) & 0.403  \\
& Dementia (10000 days) & 0.320 \\
SARD & Subsequent hospital care (10000 days) & 0.312 \\
& Initial hospital care (10000 days) & 0.305 \\
& Hospital discharge day management (10000 days) & 0.304
\\\hline
\end{tabular}
\end{table*}

\begin{table*}[ht]
\centering
\caption{Top 5 Correlated features for an example neuron from the penultimate layer of SARD models trained on the EoL task.}
\label{table:corrB}
\begin{tabular}{m{0.2\linewidth}m{0.6\linewidth}m{0.1\linewidth}}\hline
Model & Feature Name & Correlation \\\hline\hline
 & Subsequent hospital care (10000 days) & 0.417  \\
& Coronary atherosclerosis  (10000 days) & 0.399 \\
SARD (no RD) & Initial hospital care (10000 days) & 0.392 \\
& Radiologic examination, chest; single view, frontal  (10000 days) & 0.376 \\
& Hospital discharge day management  (10000 days) & 0.359
\\\hline
 & Coronary atherosclerosis (10000 days) & 0.556  \\
& Coronary arteriosclerosis in native artery (180 days) & 0.498 \\
RD Only & Congestive heart failure  (10000 days) & 0.489 \\
& Radiologic examination, chest; single view, frontal (10000 days) & 0.449 \\
& Mitral valve disorder (10000 days) & 0.439 
\\\hline
& Coronary atherosclerosis (10000 days) & 0.552  \\
& Coronary arteriosclerosis in native artery (180 days) & 0.457 \\
SARD & Congestive heart failure (10000 days) & 0.453 \\
& Mitral valve disorder (10000 days) & 0.422 \\
& Radiologic examination, chest; single view, frontal  (10000 days) & 0.385 
\\\hline
\end{tabular}
\end{table*}

\section{Cohort Inclusion Criteria and Additional Dataset Details}\label{app:dataset}

In order to ensure that patients in our dataset had sufficient medical records to learn from, we created a \emph{cohort} of patients whose medical history was sufficiently detailed for us to feel confident in making a data-driven prediction. Our inclusion criterion was that patients are enrolled in a Medicare insurance plan for all of the days in the one-year period leading up to the prediction date. For de-identification purposes, all patients whose age is over 90 have their age set to 90.

\begin{figure}[h]
    \centering
    \includegraphics[scale=0.5]{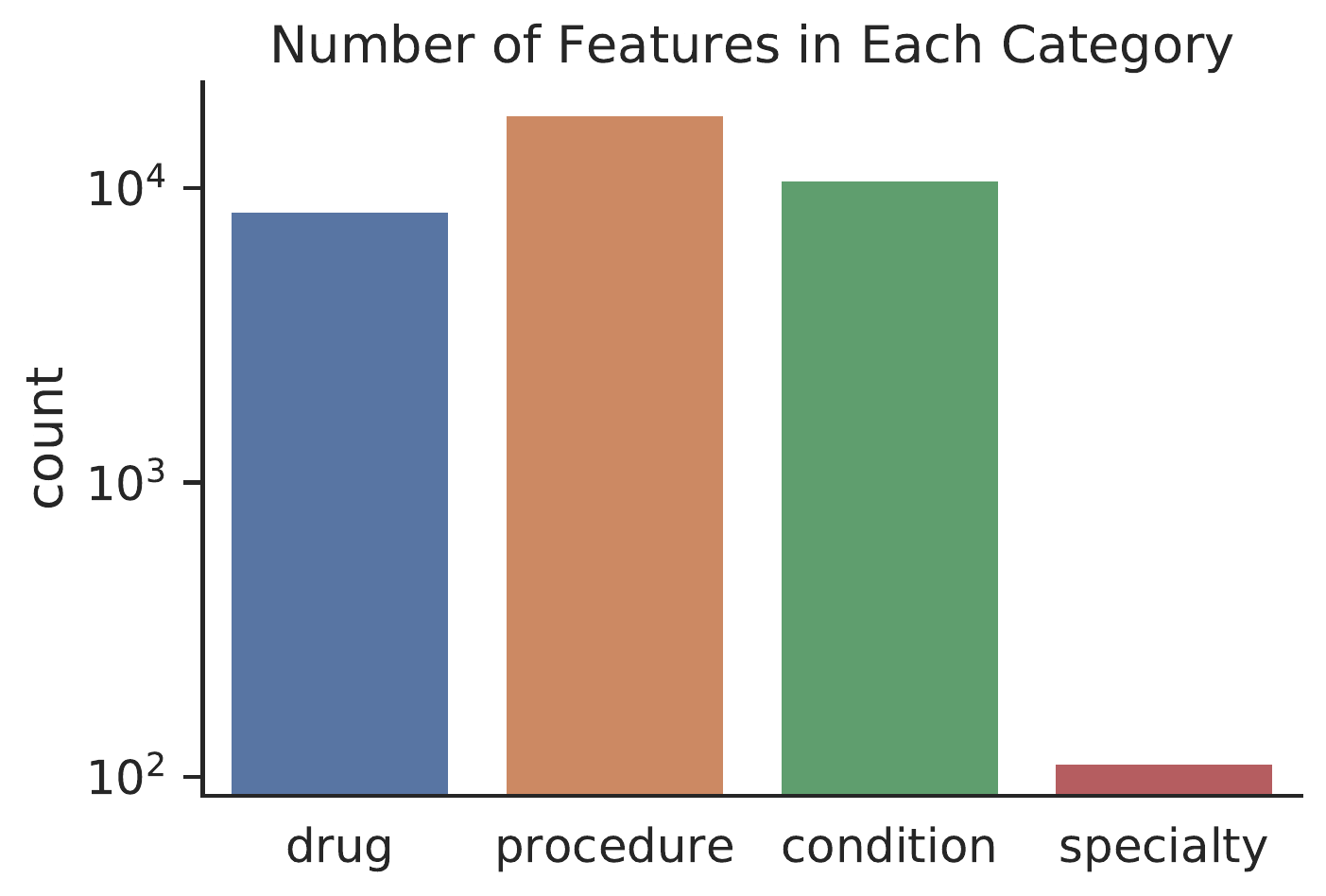}
    \caption{Breakdown of the 37,004 features in our dataset into their umbrella categories: drug administered, procedure performed, condition recorded, or specialty encountered. During a given visit, a patient will have features present from one or more of these categories.}
    \label{fig:feat_type_counts}
\end{figure}

We split the $121,593$ patients who satisfy these criteria into training, validation, and test sets of size $97,274$; $5,000$; and $19,319$ respectively. Data was collected up to the end of the calendar year 2016, with outcomes measured between April and September 2017 and a prediction date of January 1, 2017. We denote the set of all OMOP concepts used in the dataset by $\mathcal{C}$, which in our case is of size $|\mathcal{C}| = 37,004$. These codes include the drugs and procedures administered to the patient, the diagnosis codes recorded to justify these administrations, and the types of medical specialists with whom the patient interacted. The number of unique observed codes of each type in our feature set is shown in Figure \ref{fig:feat_type_counts}, indicating the rich diversity of medical information that we can utilize in our models. 

\begin{figure}[h]
  \centering
    \includegraphics[scale=0.45]{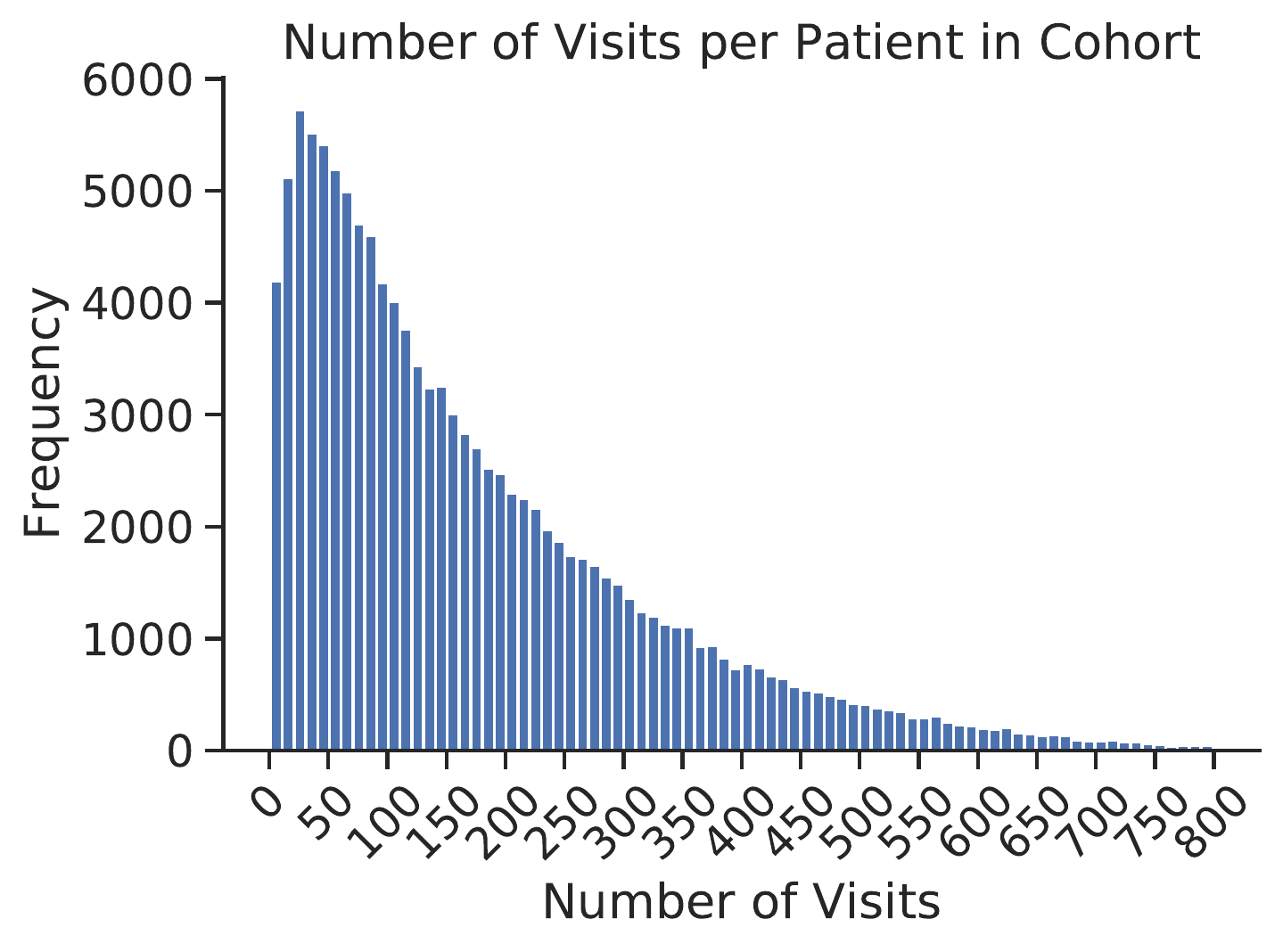}
    \caption{Histogram of the number of visits per patient. We clip the histogram at 800 visits, though a small subset of patients ($0.4\%$) have more visits. Histogram buckets have a width of 10 visits.}
    \label{fig:amt_data:a}
\end{figure}

\begin{figure}[h]
  \centering
  \includegraphics[scale=0.45]{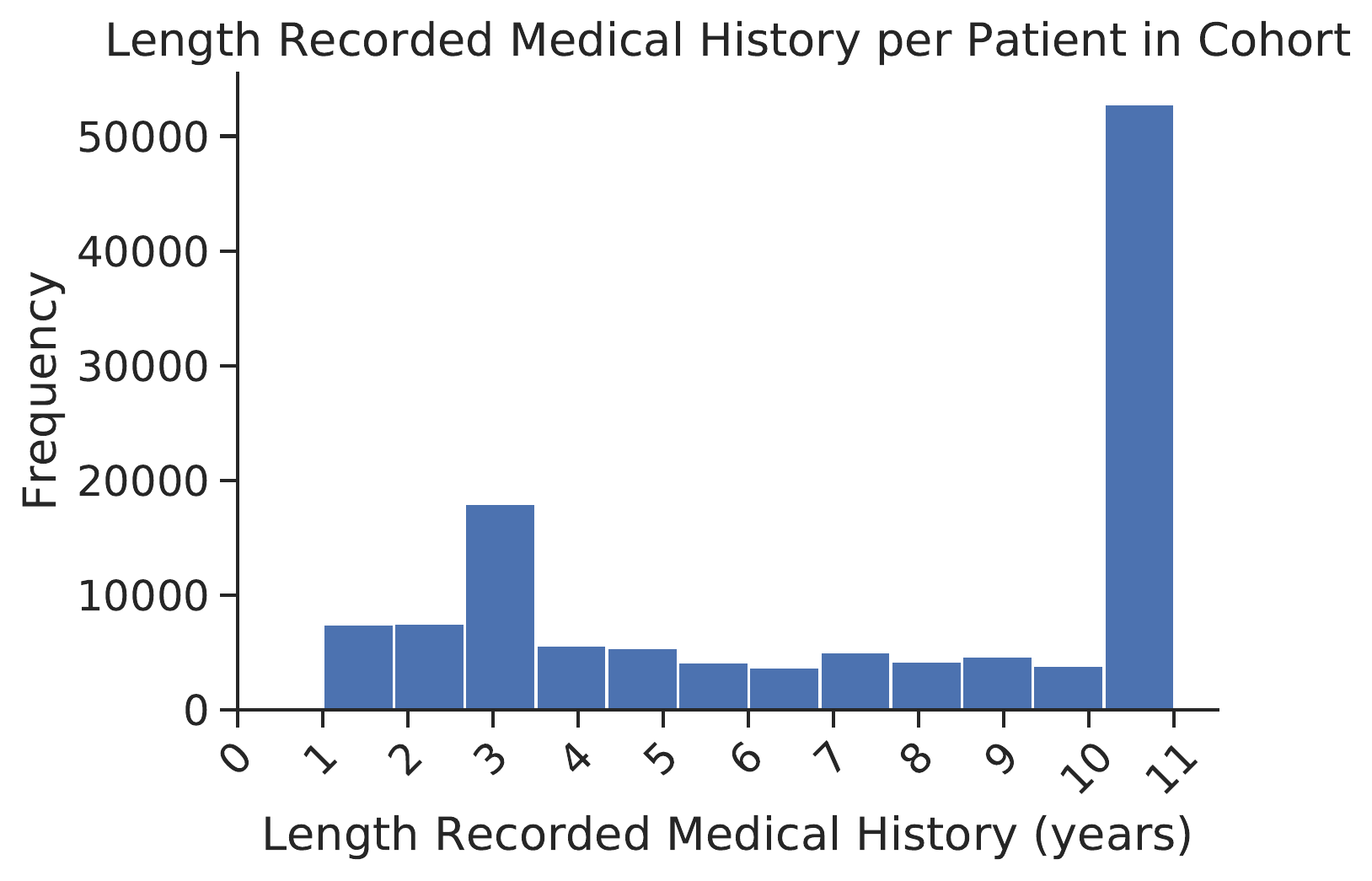}
  \caption{Histogram of recorded medical history length per patient.}
  \label{fig:amt_data:b}
\end{figure}

We receive varying amounts of data per patient, and note that the amount of data a patient has is in itself an interesting indicator of health; for example, a patient with a long medical history with very few visits may be inferred to be in better health, as they require less medical attention. We quantify the distributions characterizing the amount of information we have per patient in Figures \ref{fig:amt_data:a} and \ref{fig:amt_data:b}. As shown in Figure \ref{fig:amt_data:a}, the length of a patient's history, as measured from the time of their enrollment into an insurance plan tracked by our dataset to prediction time, ranges from $1$ to $11$ years, with a mean of $7.4$ years -- the minimum of one year results from the explicit exclusion of patients who entered a tracked insurance plan within one year of the prediction date. Additionally, as shown in Figure \ref{fig:amt_data:b}, the number of visits, or unique days during which an interaction with the healthcare system took  place, ranges from $1$ to $1,616$ per patient with a mean of $175$. We note that after embedding visits, we truncate a patient's history to include only her $512$ most recent visits. This is done for computational efficiency purposes; even after this truncation, $95.78 \%$ of patients have their complete medical history included in the dataset. 

\begin{figure}[h]
  \centering
    \includegraphics[scale=0.45]{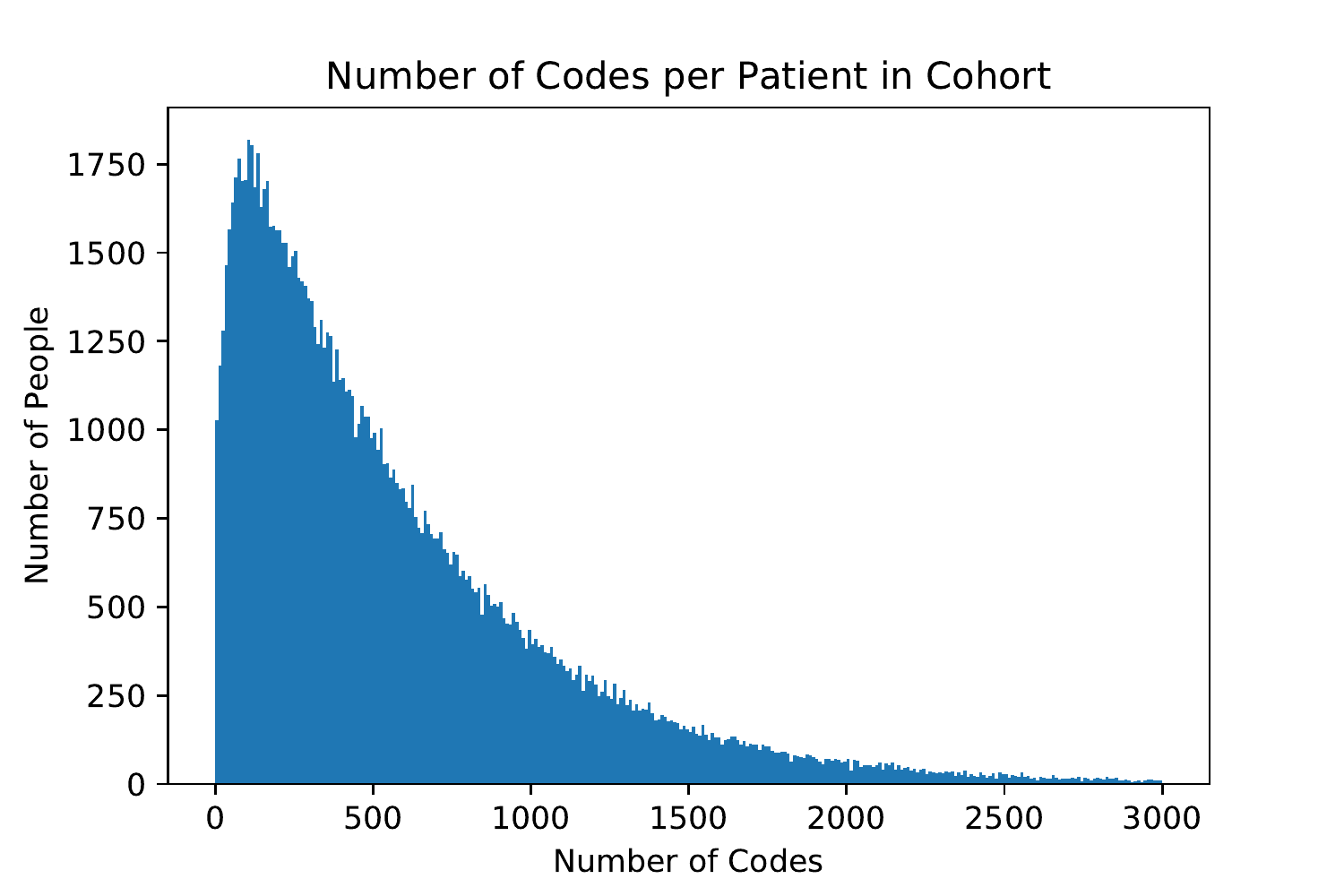}
    \caption{Histogram of the number of codes per patient. We clip the histogram at 3000 codes, though a small subset of patients ($0.4\%$) have more codes. Histogram buckets have a width of 10 codes.}
    \label{fig:codes_per_patient_loh}
\end{figure}

The BEHRT baseline uses code-level inputs instead of visit-level inputs. We quantify the amount of information we have per patient at the code level in Figure \ref{fig:codes_per_patient_loh}. Here, a patient's history is truncated to include only her 512 most recent codes, again for computational efficiency purposes; after this truncation, 57.17\% of patients have their complete medical history included in the dataset.


\end{appendices}

\end{document}